\documentclass[10pt,twocolumn,letterpaper]{article}

\usepackage[pagenumbers]{cvpr} 

\usepackage[pagebackref,breaklinks,colorlinks]{hyperref}
\usepackage{times}
\usepackage{helvet}
\usepackage{courier}
\usepackage{xcolor}
\usepackage{amssymb}
\usepackage{adjustbox}
\usepackage{subcaption}
\usepackage{graphicx} %
\usepackage{array}
\usepackage{booktabs}
\usepackage{tabularx}
\usepackage{algpseudocode}
\usepackage{algorithm}
\usepackage{amsmath}
\usepackage{siunitx}  
\usepackage{enumitem} 
\usepackage{multirow}
\usepackage{epstopdf}
\usepackage{tikz}
\usepackage{nicematrix}
\usepackage{tabularray}  
\usepackage{diagbox}

\newtheorem{proposition}{Proposition}

\usepackage{booktabs}   

\def\paperID{6191} 
\def\confName{CVPR}
\def\confYear{2026}

\title{Non-Uniform Class-Wise Coreset Selection for Vision Model Fine-tuning}

\author{
    Hanyu Zhang, Zhen Xing, Ruian He, Wenxuan Yang, Chenxi Ma, Weimin Tan, \and Bo Yan \\
    Fudan University, Shanghai, China \\
    {\tt\small hanyuzhang24@m.fudan.edu.cn}
}

\begin{document}
\maketitle
\begin{abstract}

Coreset selection aims to identify a small yet highly informative subset of data, thereby enabling more efficient model training while reducing storage overhead. Recently, this capability has been leveraged to tackle the challenges of fine-tuning large foundation models, offering a direct pathway to their efficient and practical deployment. However, most existing methods are class-agnostic, causing them to overlook significant difficulty variations among classes. This leads them to disproportionately prune samples from either overly easy or hard classes, resulting in a suboptimal allocation of the data budget that ultimately degrades the final coreset performance. To address this limitation, we propose \textbf{Non-Uniform Class-Wise Coreset Selection (NUCS)}, a novel framework that both integrates class-level and sample-level difficulty. We propose a robust metric for global class difficulty, quantified as the winsorized average of per-sample difficulty scores. Guided by this metric, our method performs a theoretically-grounded, non-uniform allocation of data selection budgets inter-class, while adaptively selecting samples intra-class with optimal difficulty ranges. Extensive experiments on a wide range of visual classification tasks demonstrate that NUCS consistently outperforms state-of-the-art methods across 10 diverse datasets and pre-trained models, achieving both superior accuracy and computational efficiency, highlighting the promise of non-uniform class-wise selection strategy for advancing the efficient fine-tuning of large foundation models.

\end{abstract}    
\section{Introduction}
\label{sec:intro}

\begin{figure}[tbp]
    \centering
    \begin{subfigure}[b]{0.49\columnwidth}
        \includegraphics[width=\textwidth]{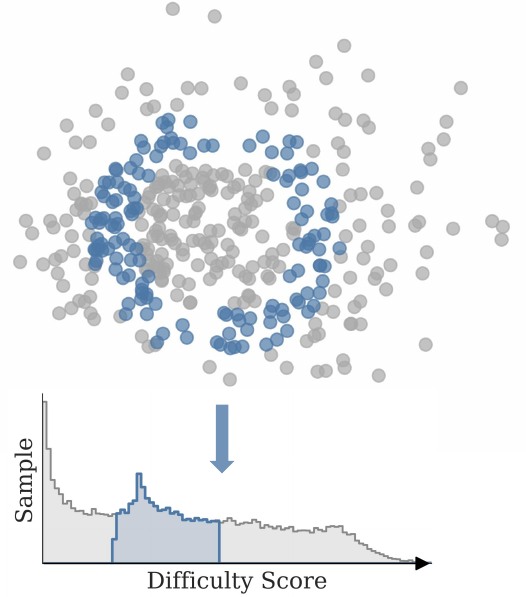}
        \caption{Class-agnostic coreset}
        \label{fig:sub_b}
    \end{subfigure}
    \hfill 
    \begin{subfigure}[b]{0.49\columnwidth}
        \includegraphics[width=\textwidth]{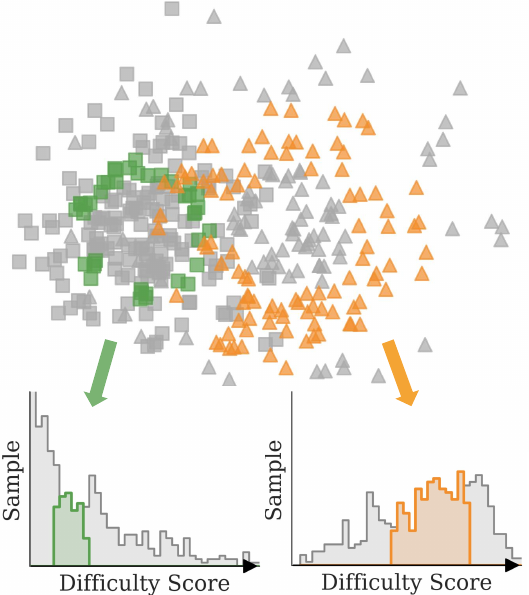}
        \caption{Non-uniform class-wise coreset}
        \label{fig:sub_a}
    \end{subfigure}    
\caption{A comparison of our method with conventional class-agnostic method. \textbf{(a)} Class-agnostic selection treats all samples as a single pool. \textbf{(b)} Our proposed NUCS performs a non-uniform, class-wise selection by choosing an appropriate number of samples within a suitable difficulty range for each class.}
    \label{fig:two_class_compare}
\end{figure}


Coreset selection aims to select a compact yet representative subset of the training data that can later be used to train future models while retaining high accuracy ~\cite{chen2010super,mirzasoleiman2020coresets,pooladzandi2022adaptive,yang2023towards}. By operating on a reduced dataset, this approach offers benefits in both storage cost and training efficiency. The relevance of this efficiency-driven paradigm has been greatly amplified with the recent advent of large-scale pre-trained foundation models. While these models have achieved unprecedented success across numerous domains through their powerful representational capabilities ~\cite{vit,isik2024scaling}, the computational and storage requirements for fine-tuning them on voluminous target datasets remain prohibitively expensive. Consequently, coreset selection has recently emerged as a compelling strategy to mitigate these burdens, with a growing body of work exploring its application in fine-tuning foundation models across various domains \cite{sorscher,staff,data,ji2025confounder, bi2025prism, das-khetan-2024-deft}.

State-of-the-art coreset selection methods are predominantly characterized by a two-stage pipeline: (1) assigning a difficulty score to each sample, and (2) subsequently pruning the dataset based on these scores~\cite{hong2024evolution, entropy2020selection, yang2024mind, zhang2024spanning}. While effective in the training-from-scratch paradigm, these approaches typically adopt a class-agnostic selection strategy, overlooking the crucial influence of class identity on coreset performance. Such strategies operate on the implicit assumption that inter-class difficulty variations are negligible. 

This assumption, however, proves untenable in the revalent paradigm of model fine-tuning. We observe the emergence of substantial inter-class difficulty variations, a phenomenon we illustrate for visual classification in \Cref{fig:two_class_compare}. These variations stem from two primary factors: (1) the model's imbalanced pre-trained knowledge relative to downstream classes, a well-documented challenge \cite{chen2025rethinking,ramanujan2023connection, shao2023investigating}, and (2) the inherent class imbalance within the downstream data.  Consequently, these class-level difficulty discrepancies cause global pruning methods to disproportionately discard samples from either overly easy or hard classes (\Cref{motivation}). While Tsai et al. \cite{tsai2025class} recently explored class-specific pruning, their method is tailored to domain-specific data, limiting its broader applicability. Furthermore, the approach relies on a simplistic uniform selection (i.e., equal sampling per class), which fails to account for significant variations in class difficulty.

In this paper, we focus on fine-tuning for visual classification, a critical task for adapting large-scale models to downstream applications. To address the aforementioned issue of inter-class difficulty variations, we first propose to quantify the global difficulty of each class as the winsorized average of sample difficulty scores within a class. Based on this quantification, we devise a class-aware budget allocation strategy and illustrate theoretically with a toy example. This strategy assigns proportionally higher selection budgets to classes quantified as more difficult, ensuring that challenging yet informative classes receive adequate representation in the final coreset, thereby counteracting the bias of global pruning methods.


We then propose NUCS, a novel coreset selection strategy that intelligently select appropriate budget and informative samples in each class. For intra-class selection, NUCS selects a continuous difficulty-ordered data range and employs linear ridge regression to adaptively identify each class's most informative subset. We conduct extensive experiments across 5 benchmark datasets spanning multiple domains, evaluating our approach on models from the two dominant architectural paradigms: Convolutional Neural Networks (ResNet) and Vision Transformers (ViT) under a wide range of pruning rates. Our experimental results demonstrate the broad applicability of the proposed NUCS method, which achieves consistently superior performance across datasets with varying class cardinalities (from 10 to 10,000 classes) and diverse distribution patterns (including balanced and long-tailed  scenarios). In summary, our contributions are:

\begin{itemize}

\item We challenge the conventional class-agnostic paradigm for coreset selection in model fine-tuning by identifying inter-class difficulty variation as a crucial yet overlooked factor, which establishes the critical necessity of a class-wise strategy.


\item Within the class-wise paradigm, we introduce a non-uniform budget allocation strategy, driven by a winsorized average measure of global class difficulty. Through both theoretical analysis and empirical validation, we establish the superiority of our adaptive approach over naive uniform selection.

\item We propose NUCS, a coreset selection framework for vision classification model fine-tuning that automatically allocates data selection budgets per class and integrates linear ridge regression to adaptively select samples intra-class, ensuring balanced representation and informativeness.

\item Through extensive experiments across 10 diverse datasets and pretrained models, we demonstrate that our methods consistently achieves state-of-the-art performance while maintaining time efficiency across a wide range of pruning rates.
    
\end{itemize}

\section{Related Work}


A prevalent strategy in coreset selection is to first quantify the difficulty of individual data samples and then select a coreset based on these scores. This section reviews key developments in this area.

\vspace{2pt}
\noindent
\textbf{Data difficulty metrics. }To measure the learning difficulty of individual training samples, a common strategy is to utilize their training dynamics \cite{forgetting,memory,aum}. The EL2N metric \cite{el2n}, for instance, calculates this by averaging a sample's prediction error over the early training epochs.

\vspace{2pt}
\noindent
\textbf{Conventional class-agnostic methods. }Earlier methods primarily focused on selecting the most difficult examples. However, this approach leads to a significant drop in precision at high pruning rates \cite{sorscher}. To improve performance across a wider range of subset sizes, more recent approaches have explored nuanced strategies. For instance, some methods advocate for prioritizing samples of moderate difficulty \cite{moderate}, while others employ a stratified sampling strategy after filtering out the most challenging samples \cite{ccs}. Other works control the difficulty distribution of the coreset by constructing a graph over the dataset \cite{d2} or adopt a "window" selection approach, which removes the easiest and hardest data points and selects from the remaining contiguous block of samples \cite{window}. These class-agnostic techniques have recently been adopted in the model fine-tuning paradigm across various domains. \citet{sorscher} investigated the efficacy of hard-selection for fine-tuning Vision Transformers (ViTs). In other domains, adaptations of stratified sampling coreset method \cite{ccs} have been used to fine-tune Large Language Models (LLMs) \cite{staff} and recommendation systems \cite{data}. More recently, \citet{bi2025prism} proposed an unsupervised method for pre-trained language models that selects samples with the lowest redundancy scores.




\vspace{2pt}
\noindent
\textbf{Class-specific coreset selection. }A significant limitation of conventional methods is their class-agnostic nature; they perform a global selection that often overlooks class-specific data characteristics. Addressing this gap, a recent line of work has started to incorporate class-level information. \citet{diet} initiated this direction by providing a preliminary analysis of how hard-selection disproportionately affects the per-class data distribution. Methodological works have also emerged to enforce class-proportional selection. For instance, \citet{window} propose maintaining the original class proportions to ensure balanced representation. More recently, \citet{tsai2025class} introduced class-balanced variants of existing methods, motivated by finding that data difficulty often exhibits strong class-clustered patterns in domains like network intrusion detection and medical imaging.






\section{Preliminary}

\subsection{Problem Definition}
Given a pruning rate $\alpha$, a target labeled dataset $\mathcal{D} = \lbrace (\mathbf{x}_i, y_i) \rbrace_{i=1}^{N}$, and a pre-trained model parameterized by $\mathbf{\theta}_S$, one-shot coreset selection for model fine-tuning selects a training subset $\mathcal{D}^\prime$ to maximize the accuracy of the model finetuned on $\mathcal{D}^\prime$. The optimization problem can be expressed as:
\begin{equation}
\min_{\substack{ |\mathcal{D}^{\prime}| \leq (1-\alpha)|\mathcal{D}|}} \mathbb{E}_{(\mathbf{x},y)\in\mathcal{D}}\left[l(\mathbf{x}, y, \mathbf{\theta}_{\mathcal{D}^{\prime}})\right]
\end{equation}
where $l$ denotes the loss function, and $\mathbf{\theta}_{\mathcal{D}^{\prime}}$ represents the model parameters after finetuning $\mathbf{\theta}_S$ with $\mathcal{D}^\prime$.

\subsection{Motivation}
\label{motivation}

Most state-of-the-art coreset selection methods leverage data difficulty metrics for sample pruning. Recent methods have demonstrated success by globally pruning the easiest or hardest examples based on difficulty scores \cite{ccs,staff,window,moderate}, but their effectiveness is compromised in the context of model fine-tuning, where substantial inter-class difficulty variations are prevalent. 

Such variations primarily stem from two factors: (1) imbalanced knowledge in the pre-trained model regarding downstream classes, and (2) potential class imbalance in the downstream data itself. The first factor can be observed through the inherent visual complexity of classes. For example, as illustrated in \Cref{fig:two_class_compare}, samples from the 'applepie' class in Food101 consistently exhibit higher EL2N scores (harder) than those from 'edamame' (easier). This suggests that the pre-trained model's representations for the visually diverse 'applepie' class are less robust than for the more uniform 'edamame' class. The second factor is a direct consequence of data distribution. On imbalanced datasets, for instance, minority classes often pose a greater learning challenge simply due to their limited number of examples, thus receiving higher difficulty scores. 

Consequently, global pruning strategies—whether removing the easiest or hardest samples—can disproportionately prune certain classes, thereby degrading overall coreset performance. For instance, pruning the 50\% hardest data globally based on EL2N scores discards 93\% of the 'applepie' class in Food101, leaving only 7\% of its samples (see \Cref{difficultyharms} for details).

The limitations of existing class-agnostic methods motivate class-aware coreset selection. To realize this, we must address two fundamental questions: (1) how to strategically allocate the total data budget across different classes (inter-class allocation), and (2) how to effectively select samples within each class's specific budget (intra-class selection).

\begin{figure}[tbp]
    \centering
    \begin{subfigure}[b]{0.48\columnwidth}
        \includegraphics[width=\textwidth]{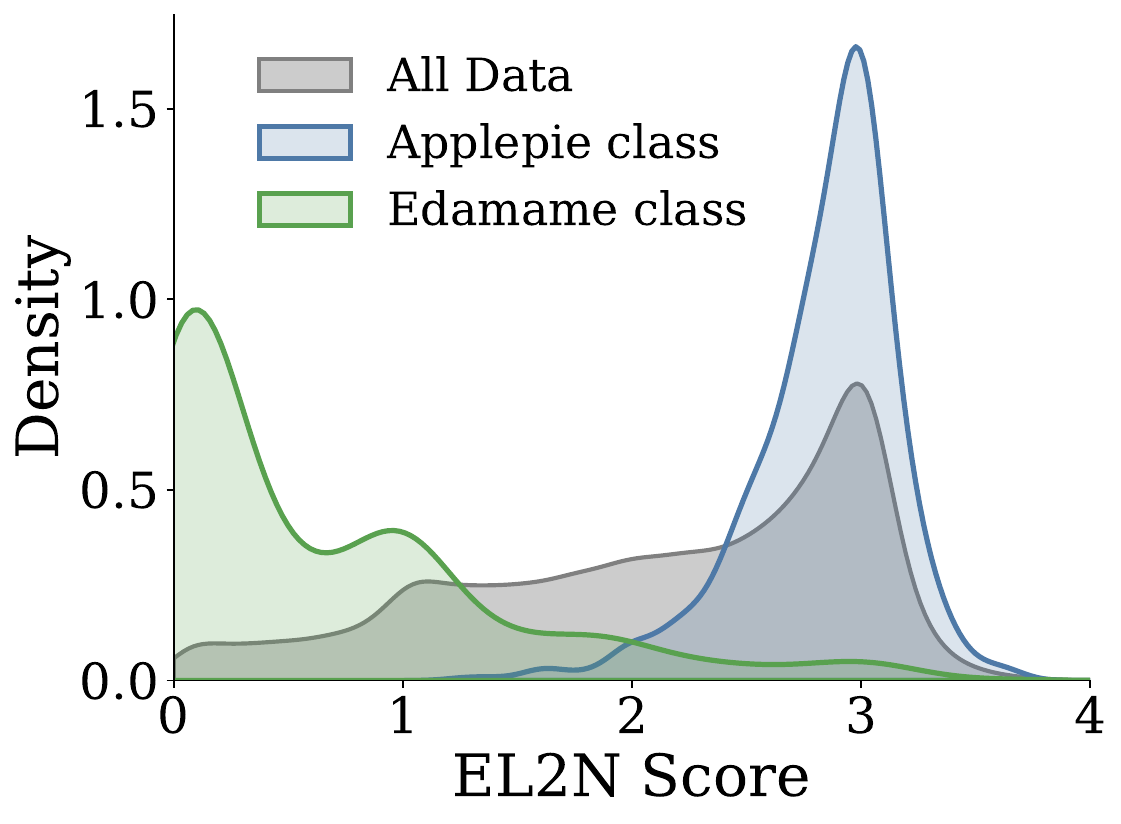}
        \caption{Food101}
        \label{fig:sub_b}
    \end{subfigure}
    \hfill 
    \begin{subfigure}[b]{0.48\columnwidth}
        \includegraphics[width=\textwidth]{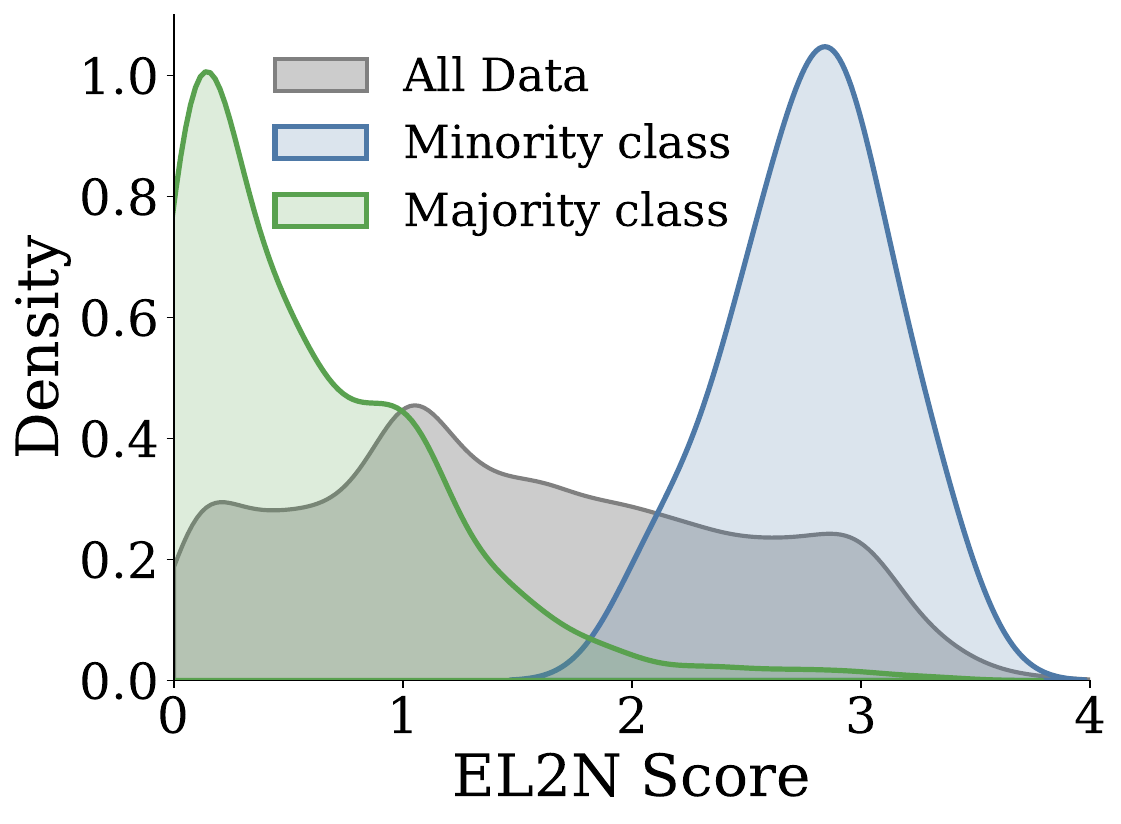}
        \caption{CIFAR100-LT}
        \label{fig:sub_a}
    \end{subfigure}
    
    \caption{
    Comparison of EL2N score distributions between the entire dataset (gray) and two individual classes (colored) on (a) Food101 and (b) CIFAR100-LT. For both datasets, the distributions of individual classes show significant shifts compared to the global data distribution, highlighting the heterogeneity of class difficulty. The model is a ResNet18 fine-tuned from ImageNet-1K.}
    \label{fig:two_class_compare}
\end{figure}



\section{Methodology }

\begin{figure*}[tbp]
    \centering
    \includegraphics[width=\textwidth]{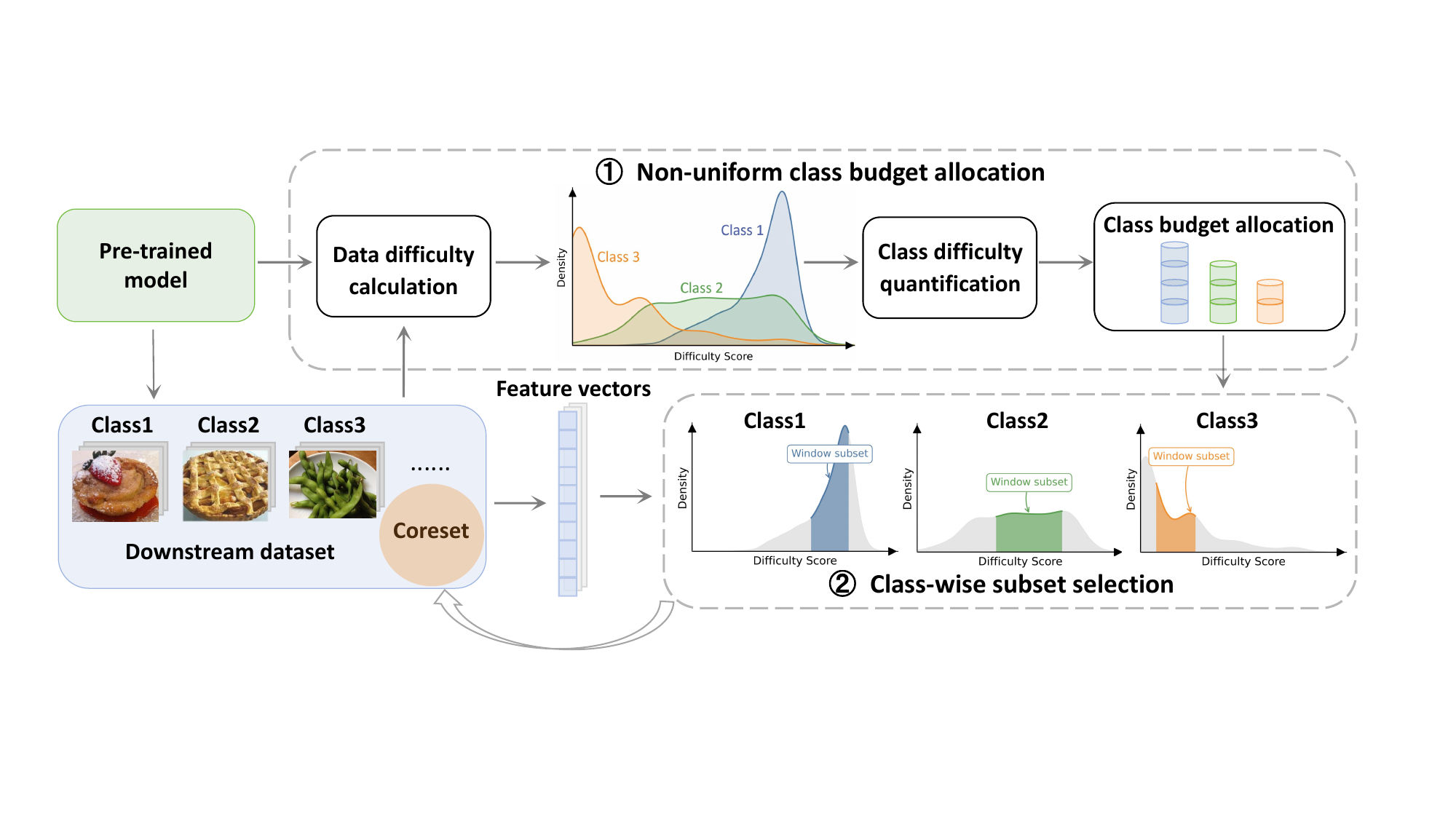}
    \caption{The overview of NUCS. In model fine-tuning, NUCS a) allocates non-uniform selection budgets based on global class difficulty b) automatically select appropriate difficulty-ranged samples in each class according to allocated budget. The workflow is illustrated here using three representative classes from the Food101 dataset, in the context of fine-tuning a ResNet18 model pre-trained on ImageNet-1K. }
    \label{fig:overview}
\end{figure*}


To perform class-aware coreset selection, in this section, we first define global class difficulty and theoretically demonstrate that non-uniform budget allocation according to global class difficulty helps (\Cref{3.2}). Building upon this theoretical insight, we propose Non-Uniform Class-Wise Coreset Selection (NUCS), a novel framework that automatically determines both the appropriate quantity and difficulty distribution of samples to select for each class (\Cref{NUCS}).

\subsection{Non-Uniform Class Budget Allocation}
\label{3.2}

\subsubsection{Quantifying Global Class Difficulty}

Our non-uniform budget allocation is motivated by the significant variation in learnability observed across downstream classes. To quantify this variation, we introduce the metric for \textit{global class difficulty} $\mathbf{S}_j$.

This metric is derived from sample-level difficulty scores, $s(x_i) \in \mathbb{R}$, assigned to each sample $x_i$ in a class $\mathcal{D}_j$, where higher values denote greater difficulty (e.g., computed via EL2N). To ensure the class difficulty estimate is robust to potential outliers, we aggregate these scores using the winsorized average. Formally, given the scores $s_{1} \le s_{2} \le \dots \le s_{N_j}$ for class $j$ sorted in ascending order and a winsorization fraction $\gamma$, the global class difficulty $\mathbf{S}_j$ is computed as:
\begin{equation}
    \mathbf{S}_j = \frac{1}{N_j} \left( \sum_{i=k+1}^{N_j-k} s_{i} + k \cdot s_{k+1} + k \cdot s_{N_j-k} \right),
    \label{eq:winsorized_avg}
\end{equation}
where $k = \lfloor \gamma N_j \rfloor$. We set $\gamma = 0.05$ for all our experiments.

To validate our initial premise, we analyzed the coefficient of variation of $\lbrace \mathbf{S}_j\rbrace$ across all classes. As reported in \Cref{cv of class}, the results reveal a substantial disparity in global class difficulty for various pre-trained models and datasets. Crucially, this disparity is markedly more pronounced than in models trained from scratch, substantiating the necessity for our proposed strategy.

\subsubsection{Difficulty-Guided Budget Allocation}
\label{sec:budget_allocation}

This metric $\mathbf{S}_j$ is the cornerstone of our allocation strategy. Our central hypothesis is that prioritizing harder classes leads to a more effective coreset. We formalize this guiding principle as a proposition:

\begin{proposition}
\label{prop:main_idea}
A non-uniform allocation strategy that assigns a larger portion of the budget to classes with higher global difficulty $\mathbf{S}_j$ yields a more effective coreset.
\end{proposition}

We instantiate this proposition with a concrete allocation strategy, detailed in \Cref{NUCS}. To demonstrate that our difficulty-guided allocation is not contingent on a specific difficulty metric, we evaluate it using three distinct options: EL2N \cite{el2n}, Effort~\cite{data}, and AUM~\cite{aum}.
Among these, EL2N is our primary metric, used for computing the sample scores $s(x_i)$ (see \Cref{el2n} for details). As detailed in \Cref{sec:ablation}, our method consistently outperforms uniform baselines across all three, validating its general applicability.

\subsubsection{Theoretical Illustration}

To provide a theoretical grounding for our guiding principle in \Cref{prop:main_idea}, we analyze a simplified model that illustrates why prioritizing more difficult classes is a principled strategy. We adapt a binary classification setup, originally from \citet{drop}, by incorporating a notion of class difficulty. 



Consider a dataset consisting of two classes, $\mathcal{D}_0$ and $ \mathcal{D}_1$, each containing $ N $ data points. Let $ f $ denote the data selection rate, and $ f_0 $ and $ f_1 $ represent the class selection rates for $\mathcal{D}_0$ and $ \mathcal{D}_1$. Here, we have $f_0 + f_1 = 2f$. The data for each class follows an independent Gaussian distribution. We assume that after random pruning, the distribution of the selected data within each class remains Gaussian: $x_0 \sim \mathcal{N}(\mu_0, \sigma_0)$ for class $ \mathcal{D}_0$, and $x_1 \sim \mathcal{N}(\mu_1, \sigma_1)$ for class $ \mathcal{D}_1$. Without loss of generality, we assume $\mu_0 < \mu_1$. We adopt a linear decision rule, where a sample $x$ is classified as $ \mathcal{D}_0$ if $x \leq t$ and as $ \mathcal{D}_1$ if $x > t$.  For a sufficiently large $N$, the empirical distribution of the selected subset within each class will closely approximate the assumed Gaussian distribution. Consequently, the classification error rates can be estimated using the standard normal cumulative distribution function $\Phi$ as:
\begin{equation}
    E_{0}(t) = \Phi\left(\frac{\mu_0-t}{\sigma_0}\right), \quad E_1(t) = \Phi\left(\frac{t-\mu_1}{\sigma_1}\right).
\end{equation}
The overall error rate is:
\begin{equation}
    E(t,f_0) = f_0 \Phi\left(\frac{\mu_0-t}{\sigma_0}\right)+(2f-f_0)\Phi\left(\frac{t-\mu_1}{\sigma_1}\right).
\end{equation}
The optimal decision boundary $t$ and selection rate $f_0$ should minimize this error. By setting the partial derivatives to zero, we immediately obtain:
\begin{equation}
\begin{cases}
    \Phi\left(\frac{\mu_0-t}{\sigma_0}\right) = \Phi\left(\frac{t-\mu_1}{\sigma_1}\right) \\
    \frac{f_0}{\sigma_0} \Phi'\left(\frac{\mu_0-t}{\sigma_0}\right)= \frac{f_1}{\sigma_1} \Phi'\left(\frac{t-\mu_1}{\sigma_1}\right)
\end{cases}
\end{equation}
Solving this system of equations yields the optimal decision boundary $t$ and the relationship between $f_0$ and $f_1$:
\begin{align}
    t &= \frac{\sigma_1 \mu_0 + \sigma_0 \mu_1}{\sigma_0 + \sigma_1}, \\
    f_0 &= \frac{\sigma_0}{\sigma_1} f_1.
\end{align}

In this idealized setting, the class variance $\sigma_i$ serves as a measure of difficulty. A larger $\sigma_i$ implies a more dispersed distribution, leading to greater overlap with the other class and a higher error rate for any linear ecision boundary. The optimal boundary $t^*$ equalizes the error for both classes, $E_0(t^*) = E_1(t^*)$, and this equilibrium error increases with either $\sigma_0$ or $\sigma_1$. 


The optimal allocation derived from our model, $f_i \propto \sigma_i$, dictates that the budget allocated to a class should be proportional to its difficulty.  While we do not expect this exact linear relationship to hold in the complex of deep networks, this analysis provides a theoretical rationale for our core hypothesis.

\subsection{NUCS}
\label{NUCS}

\begin{algorithm}[t]
\caption{NUCS: Non-Uniform Class-Wise Coreset Selection}
\label{alg:nucs}
\begin{algorithmic}[1]
\Statex \hspace{-1.8em} \textbf{Input:} 
    Pre-trained model with feature extractor $F(\cdot)$;
    downstream dataset $\mathcal{D}=\lbrace (\mathbf{x}_i,y_i)\rbrace_{i=1}^{N}$ with $Y$ classes; pruning rate $\alpha$; data difficulty scores $\{s_i\}_{i=1}^N$; window step size t.
\State Extract $\mathcal{D}_F \leftarrow \lbrace (\mathbf{F}_i,y_i)|\mathbf{F}_i=F(\mathbf{x}_i)\rbrace_{i=1}^{N}$.

\State Compute global class difficulties $\lbrace \mathbf{S}_j \rbrace_{j=1}^{Y}$ according to \Cref{eq:winsorized_avg}.

\For{each class $j=1, \dots, Y$ with $N_j$ samples}
    \State Calculate class selection budget: $b_j\leftarrow \min (\lfloor  \frac{(1-\alpha) \mathbf{S_j} N_j N}{\sum_{j'=1}^{Y}\mathbf{S}_{j'}N_{j'}}  \rfloor ,N_j)$.
\EndFor
\State Allocate remaining budget if $\sum_{i=1}^{Y}b_j<N$.
\For{$k \in \lbrace 0, t, 2t, \dots, 1 \rbrace$} 
    \State Initialize candidate coreset $C_k\leftarrow \varnothing$.
    \For{each class $j=1, \dots, Y$}
        \State Sort samples of class $j$ by difficulty score $s_i$ in ascending order.
        \State Define window start and end: $i_{start} \leftarrow \lfloor k \cdot N_j \rfloor - b_j$, $i_{end} \leftarrow \lfloor k \cdot N_j \rfloor$.
        \State $C_{j,k} \leftarrow$ Select samples from index $i_{start}$ to $i_{end}$ in the sorted list.
        \State $C_k\leftarrow C_k\cup C_{j,k}$.
    \EndFor
    \State Train linear ridge regression on $\lbrace(\mathbf{F}_i, y_i) | (\mathbf{x}_i, y_i) \in C_k \rbrace$ to get weights $\mathbf{w}_k$.
    \State Evaluate $\mathbf{w}_k$ on the full feature-label set $\mathcal{D}_F$ to get validation performance $P_k$.
\EndFor
\State Find optimal window: $k^* = \arg\max_{k} P_k$.
\State \Return $C_{k^*}$.
\end{algorithmic}
\end{algorithm}

Given a pre-trained model with feature extractor $F$, a target downstream dataset $\mathbf{D} = \{(x_i, y_i)\}_{i=1}^{N}$ for $Y$-way classification and a pruning rate $\alpha$, we define $N_j$ as the initial sample count for class $j$, so $N=\sum_{j=1}^{Y}{N_j}$. The proposed Non-Uniform Class-Wise Coreset Selection framework is designed to strategically select both the appropriate selection rate and representative data for each class in $\mathbf{D}$. 

\vspace{2pt}
\noindent
\textbf{Non-uniform class budget allocation. }First, we fine-tune the pre-trained model on the downstream dataset $\mathbf{D}$ for a few epochs and compute the EL2N scores $\lbrace s_{i}\rbrace_{i=1}^{N}$. These scores are then aggregated to obtain the global class difficulty $s_j$ for each class according to \Cref{eq:winsorized_avg}. The data selection budget $b_j$ for class $j$ is set as $b_j= \frac{(1-\alpha)\mathbf{S_j} N_j}{T} $, where $T=\frac{\sum_{i=1}^{Y}\mathbf{S_i}N_i}{N}$ serves as a normalization factor that maintains the global pruning rate $\alpha$. To ensure that the number of selected samples does not exceed the available data, we cap the value of $b_j$ at $N_j$. Specifically, if $b_j > N_j$, we set $b_j = N_j$, utilizing all samples from that class and redistribute the remaining budget to other classes according to their budget proportions. This formulation ensures that the data selection rate for each class is positively correlated with its relative difficulty, thereby giving greater attention to more challenging classes.

\begin{figure*}[tbp]
    \centering

    \begin{subfigure}[b]{0.32\textwidth}
        \includegraphics[width=\linewidth]{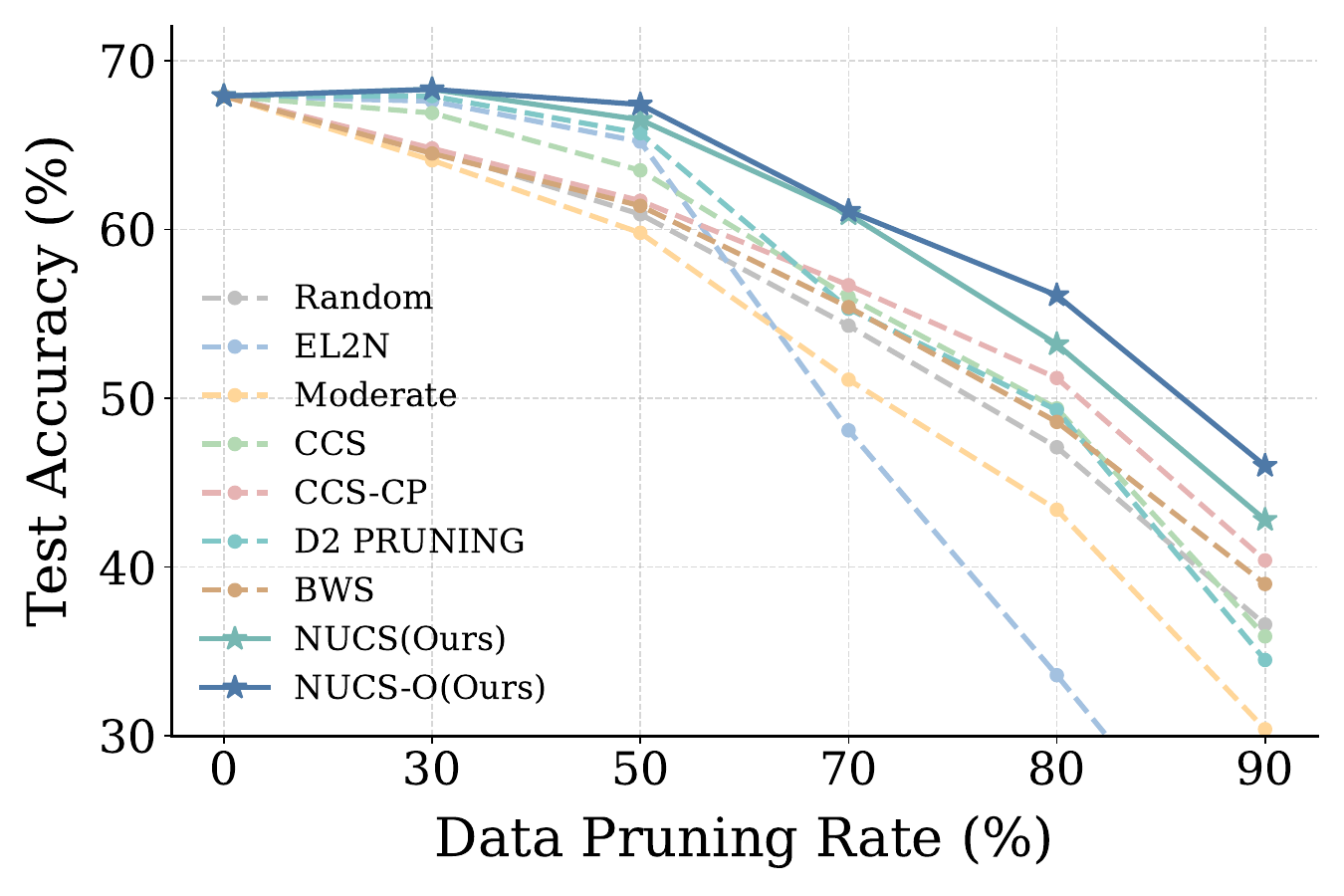}
        \caption{CIFAR100-LT \\ ResNet18 (IN-1K)}
        \label{fig:sub_lt_cifar100_resnet} 
    \end{subfigure}
    \hfill 
    \begin{subfigure}[b]{0.32\textwidth}
        \includegraphics[width=\linewidth]{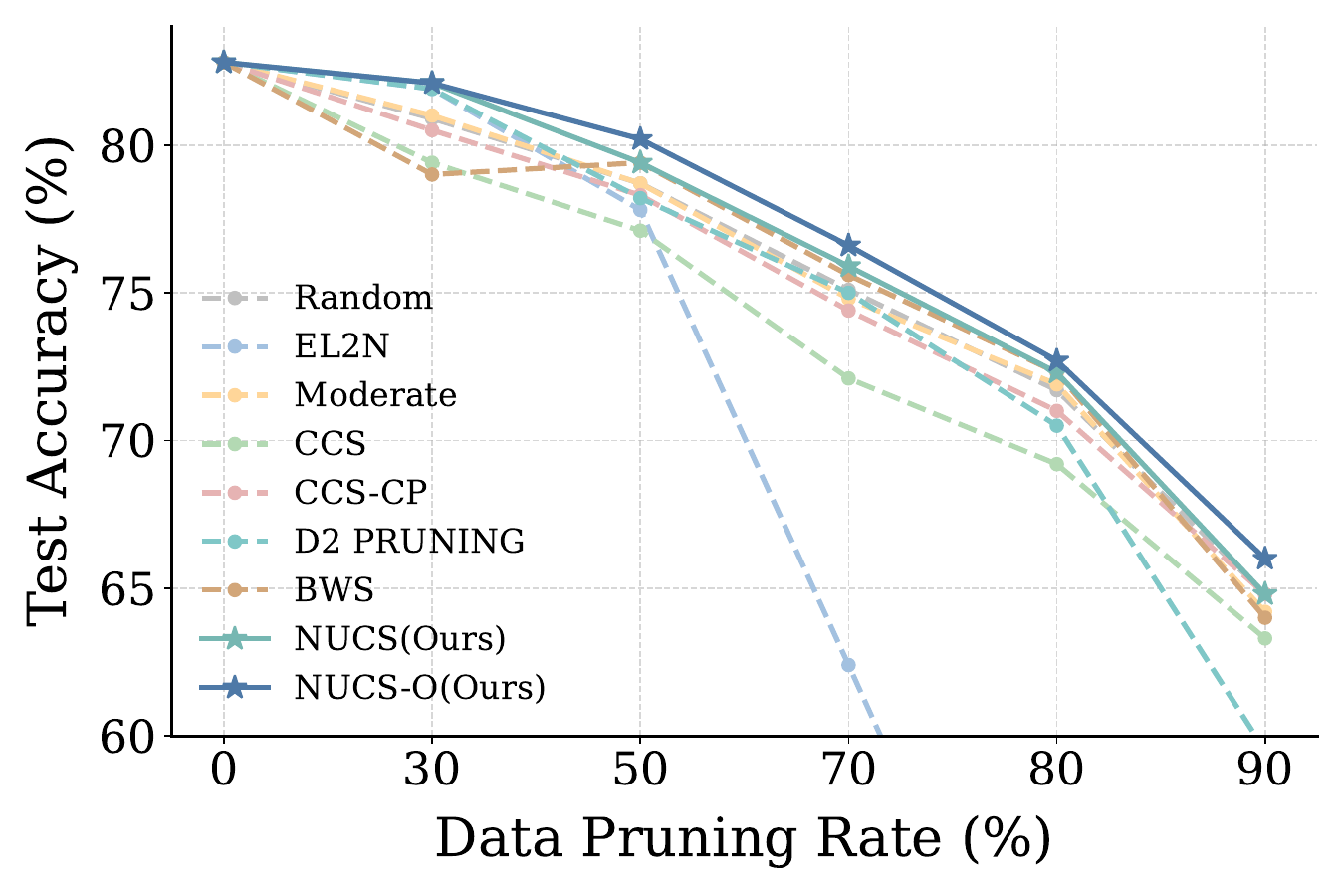}
        \caption{Food101 \\ ResNet18 (IN-1K)}
        \label{fig:sub_food101_resnet}
    \end{subfigure}
    \hfill 
    \begin{subfigure}[b]{0.32\textwidth}
        \includegraphics[width=\linewidth]{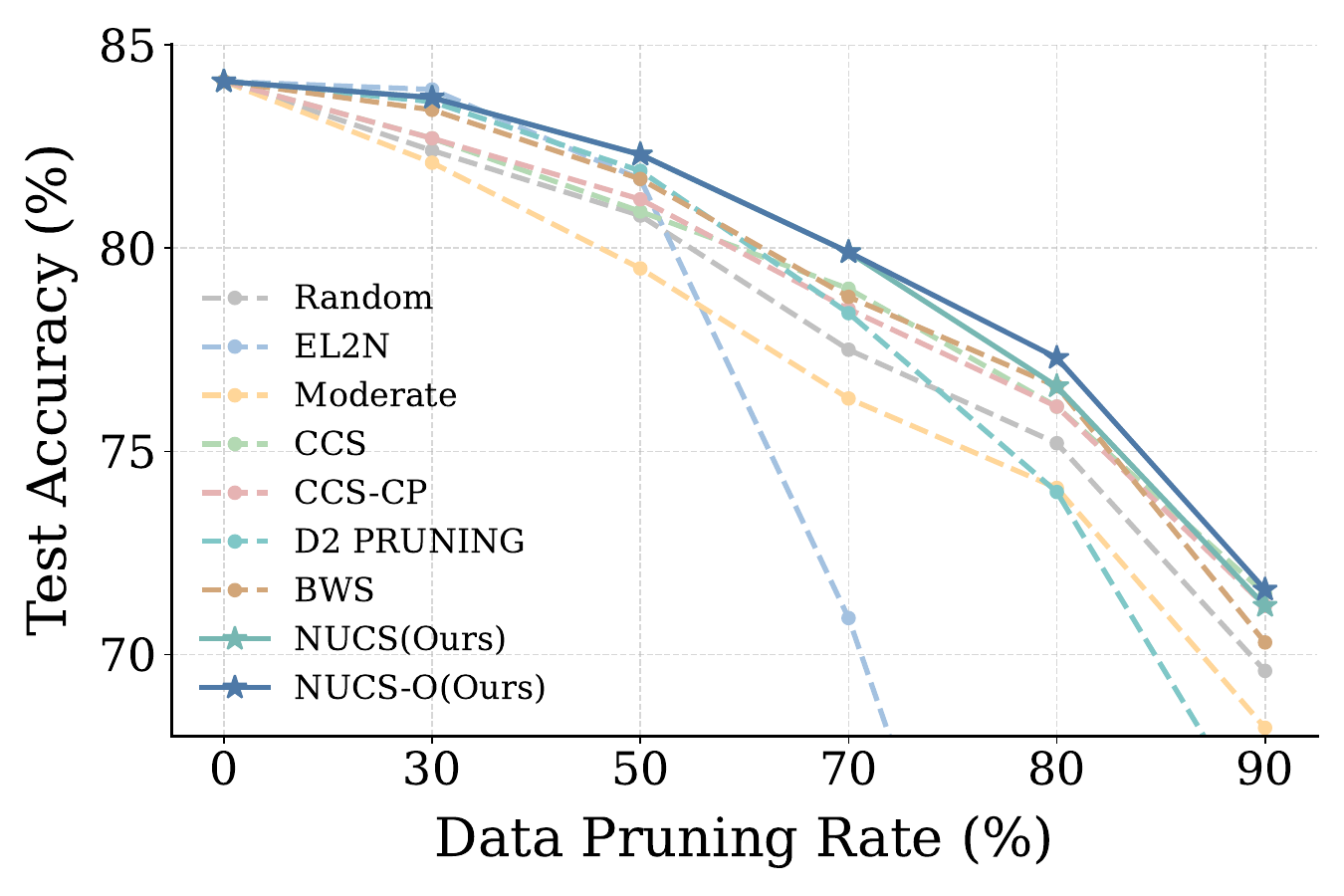}
        \caption{CIFAR100 \\ ResNet18 (IN-1K)}
        \label{fig:sub_cifar100_resnet}
    \end{subfigure}

    \vspace{-0.5pt} 

    \begin{subfigure}[b]{0.32\textwidth}
        \includegraphics[width=\linewidth]{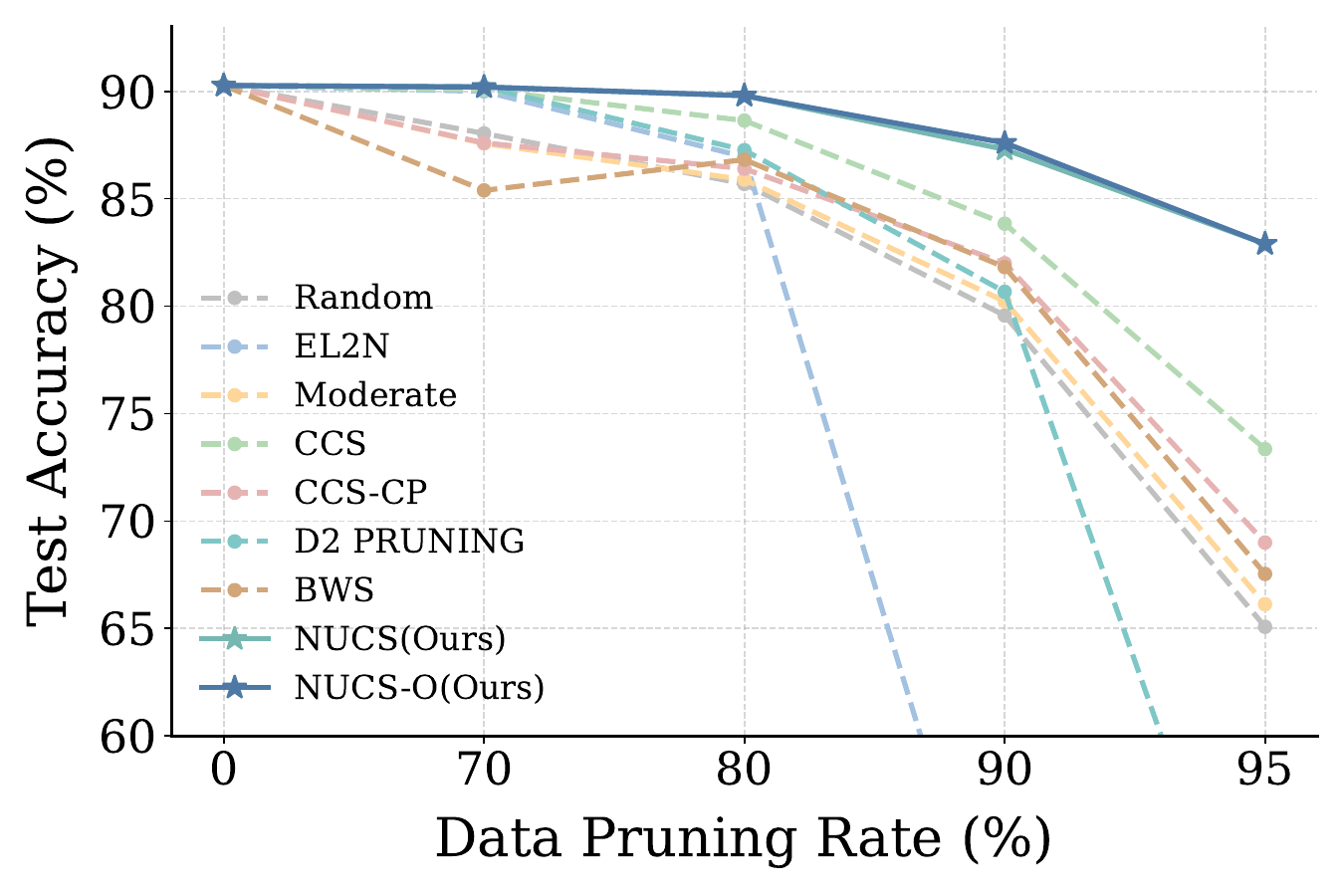}
        \caption{CIFAR100-LT \\ ViT-L (IN-21K)}
        \label{fig:sub_lt_cifar100_vit}
    \end{subfigure}
    \hfill 
    \begin{subfigure}[b]{0.32\textwidth}
        \includegraphics[width=\linewidth]{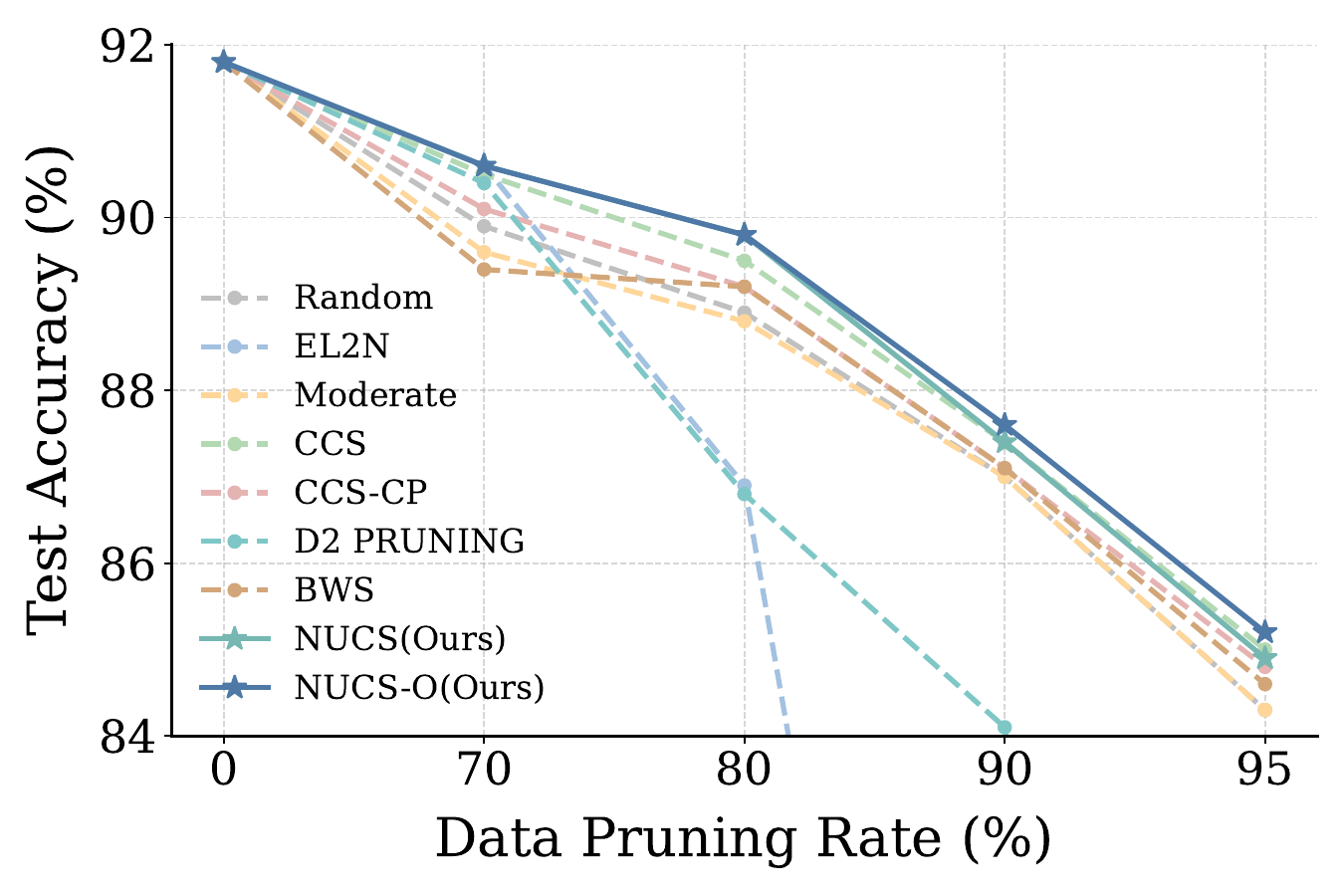}
        \caption{Food101 \\ ViT-L (IN-21K)}
        \label{fig:sub_food101_vit}
    \end{subfigure}
    \hfill
    \begin{subfigure}[b]{0.32\textwidth}
        \includegraphics[width=\linewidth]{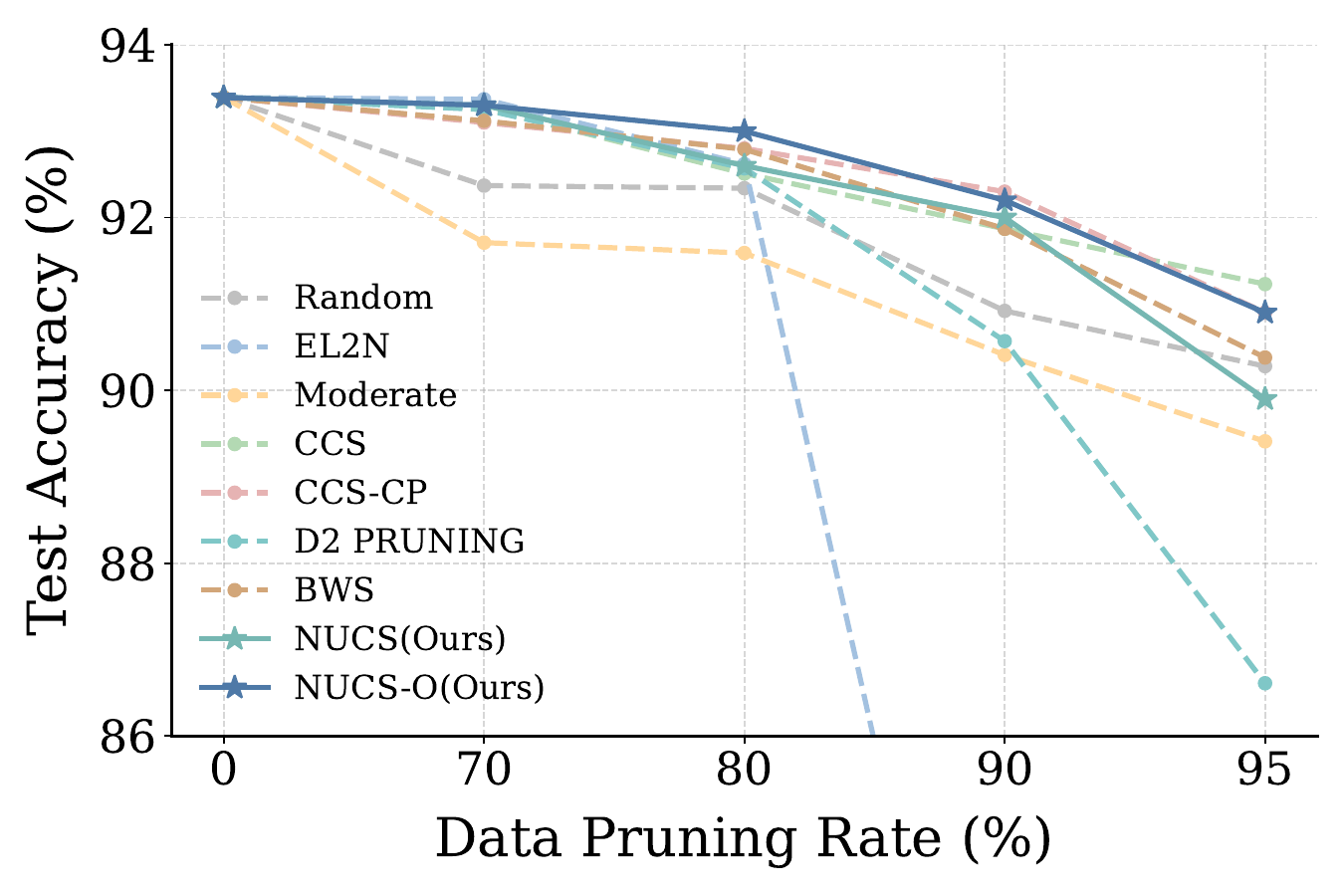}
        \caption{CIFAR100 \\ ViT-L (IN-21K)}
        \label{fig:sub_cifar100_vit}
    \end{subfigure}
    \caption{Performance comparison between our methods and other baselines. Experimental results demonstrate consistent and significant improvements across various datasets and pre-trained models. To account for the superior robustness of pretrained ViT model against data pruning, we evaluate them under a more challenging set of higher pruning rates.}
    \label{fig:pruning_results_combined}
\end{figure*}

\vspace{2pt}
\noindent
\textbf{Window-based intra-class coreset selection. }We perform coreset selection on a per-class basis using the windowing method of \citet{window}. 
For each class~$j$, we first sort its $N_j$ samples in ascending order based on their difficulty scores. Given a class-specific budget $b_j$, we select a coreset corresponding to a continuous block of indices. This window is defined by the interval $[\lfloor k \cdot N_j \rfloor - b_j, \lfloor k \cdot N_j \rfloor]$, where $k \in [0, 1]$ is a global hyperparameter that sets the window's endpoint as a fraction of the class size. If the start index $\lfloor k \cdot N_j \rfloor - b_j$ is negative, the selection window is adjusted to $[0, b_j]$.


\vspace{2pt}
\noindent
\textbf{Optimal window subset determination. }The optimal difficulty profile of a coreset depends on the pruning rate~\cite{ccs}. Consequently, the optimal window endpoint $k$ should be adjusted according to the desired coreset size. We denote the coreset found via grid search of k as NUCS-O, which serves as our performance upper bound. To circumvent this prohibitive grid search, we propose to directly predict the optimal endpoint $k$ using linear ridge regression. This approach is motivated by the demonstrated success of regression-based methods in analogous data pruning tasks \cite{window,lee2024selfsupdd}. Specifically, we use the feature extractor $F$ to obtain an embedding vector $\mathbf{F_i} = F(x_i)$ for each sample. For each candidate window subset $C_k$, we construct a feature matrix $\mathbf{X}_k$ by stacking the corresponding feature vectors $\{\mathbf{F_i}\}_{i \in I_k}$ as rows, and a label vector $\mathbf{y}_k$. The ridge regression problem for this subset is formulated to find the optimal weight vector $\mathbf{w}_k$:
\begin{equation}
    \mathbf{w}_k := \arg\min_{\mathbf{w}} \|\mathbf{y}_k - \mathbf{X}_k \mathbf{w}\|_2^2 + \lambda\|\mathbf{w}\|_2^2.
\end{equation}
After solving for $\mathbf{w}_k$ for each candidate window, we evaluate its classification performance on the entire training set. The window subset $C_k$ whose corresponding model $\mathbf{w}_k$ achieves the highest validation accuracy is selected as the final coreset.




\section{Evaluation}

\textbf{Datasets and Models. }To comprehensively evaluate NUCS, we conducted experiments on four balanced datasets (CIFAR10~\cite{cifar}, CIFAR100, Food101~\cite{food}, iNaturalist 2021 mini~\cite{inaturalist2021}) and CIFAR100-LT (a long-tailed version of CIFAR 100) with imbalance factor I=20. Our evaluation framework employed two pre-trained models: ResNet18~\cite{resnet} (on ImageNet-1K) and ViT-L~\cite{vit} (on ImageNet-21K).

\vspace{2pt}
\noindent
\textbf{Baselines. }We compare NUCS against two types of coreset selection baselines:
\begin{itemize}
    \setlength{\itemsep}{0pt} 
    \setlength{\parsep}{0pt}

    \item \textbf{Class-Agnostic Methods.} These approaches select data from the entire dataset without explicitly enforcing per-class selection quotas.
    \begin{itemize}
        \setlength{\itemsep}{0pt}
        \setlength{\parsep}{0pt}
        \item Random selection.
        \item EL2N~\cite{el2n} Selects samples with the highest error L2-norm scores.
        \item Moderate~\cite{moderate} Selects samples of intermediate difficulty based on feature-space distances.
        \item CCS~\cite{ccs} Constructs the coreset via stratified sampling on sample scores.
        \item D2 PRUNING~\cite{d2} Models the dataset as a graph to balance sample difficulty and diversity.
    \end{itemize}

    \item \textbf{Class-Specific Methods. }These methods explicitly maintain the proportion of data in each class during selection.
    \begin{itemize}
        \setlength{\itemsep}{0pt}
        \setlength{\parsep}{0pt}
        \item BWS~\cite{window} Selects the coreset from difficulty-sorted intervals within each class.
        \item CCS-CP~\cite{tsai2025class} The class-proportional variant of CCS that preserves the original class distribution.
    \end{itemize}
\end{itemize}


\vspace{2pt}
\noindent
\textbf{Implementation. }The data difficulty is quantified using the EL2N score. The window step size $t$ for NUCS and NUCS-O is set as 0.1. For model fine-tuning, the final pre-trained layer is replaced with a linear classifier. The regularization parameter $\lambda$ in linear ridge regression is set to 1. The long-tailed datasets are created following~\citet{classbalance}. We provide more training details in \Cref{app:setting}.

\subsection{Coreset Performance Comparsion}

We compare NUCS with other coreset selection methods on both balanced and imbalanced datasets. Notably, some state-of-the-art methods such as CCS-CP require hyperparameter grid search. To ensure a fair comparison, we also provide the results of NUCS-O, where the class window subset fraction endpoint k is determined through grid search.

\vspace{2pt}
\noindent
\textbf{Main results. }Both NUCS and NUCS-O perform well across a wide range of pruning rates and pre-trained models, with NUCS achieving highly competitive results and NUCS-O demonstrating the best overall performance (full results in \Cref{fig:pruning_results_combined}). For example, at a 90\% pruning rate with ResNet18 (IN-1K), NUCS-O achieves significant improvements over the state-of-the-art methods: 1.2\% on Food101 and 5.6\% on CIFAR100-LT, while our base method NUCS concurrently secures the second-best position, outperforming all other competitors.

\begin{table}[tbp]
\centering
\sisetup{
    table-format=2.1,
    detect-weight,
    mode=text
}
\setlength{\tabcolsep}{4pt} 

\caption{
 Performance vs. hyperparameter-free baselines on iNaturalist 2021 mini dataset (500K data, 10,000 classes).
}
\label{tab:inat}
\renewcommand{\arraystretch}{0.9} 

\begin{NiceTabular}{
    l l                  
    S
    S
    S
    S
    S
}
\CodeBefore
  \rowcolor{gray!20}{3,7} 
\Body
\toprule
& & \multicolumn{5}{c}{Data Pruning Rate } \\ 
\cmidrule(lr){3-7}
\textbf{Model} & \textbf{Method} & {0\%} & {20\%} & {60\%} & {70\%} & {80\%} \\
\midrule
\multirow{4}{*}{ResNet18} 
& Random          & 44.1 & 39.9 & 26.8 & 21.7 & 15.4  \\
& Moderate        & {-}  & 38.9 & 24.1 & 19.6 & 14.3  \\
& CCS             & {-}  & \bfseries 40.1 & 26.5 & 20.9 & 14.0 \\ 
& NUCS            & {-}  & \bfseries 40.1 & \bfseries 27.4 & \bfseries 22.6 & \bfseries 16.1  \\
\midrule
\multirow{4}{*}{ViT-L} 
& Random          & 58.6 & 56.8 & 49.0 & 44.3 & 36.8  \\
& Moderate        & {-}  & 55.8 & 45.8 & 41.8 & 35.4  \\
& CCS             & {-}  & 54.6 & 48.1 & 39.7 & 28.6  \\
& NUCS            & {-}  & \bfseries 58.1 & \bfseries 52.9 & \bfseries 49.5 & \bfseries 43.8  \\
\bottomrule
\end{NiceTabular}
\end{table}

\vspace{2pt}
\noindent
\textbf{Scalability analysis. }To show the scalability of our method, we evaluate NUCS with other hyperparameter-free methods on the large-scale iNaturalist 2021 mini dataset (500K samples, 10,000 classes; experimental result shown in \Cref{tab:inat}). Conventional methods like CCS face significant computational challenges due to their requirement for exhaustive hyperparameter grid searches. Following the approach of \cite{staff,data}, we implemented CCS with a hard cutoff rate $\beta=0$. Our experiments reveals that many hyperparameter-free methods fails to maintain its effectiveness and even underperform random selection. In contrast, our proposed method demonstrates remarkable scalability, maintaining its superior performance on the large-scale dataset with massive class diversity. 

\begin{figure}[t]
    \centering
    \begin{subfigure}[b]{0.48\columnwidth} 
        \centering
        \includegraphics[width=\textwidth, height=6cm, keepaspectratio]{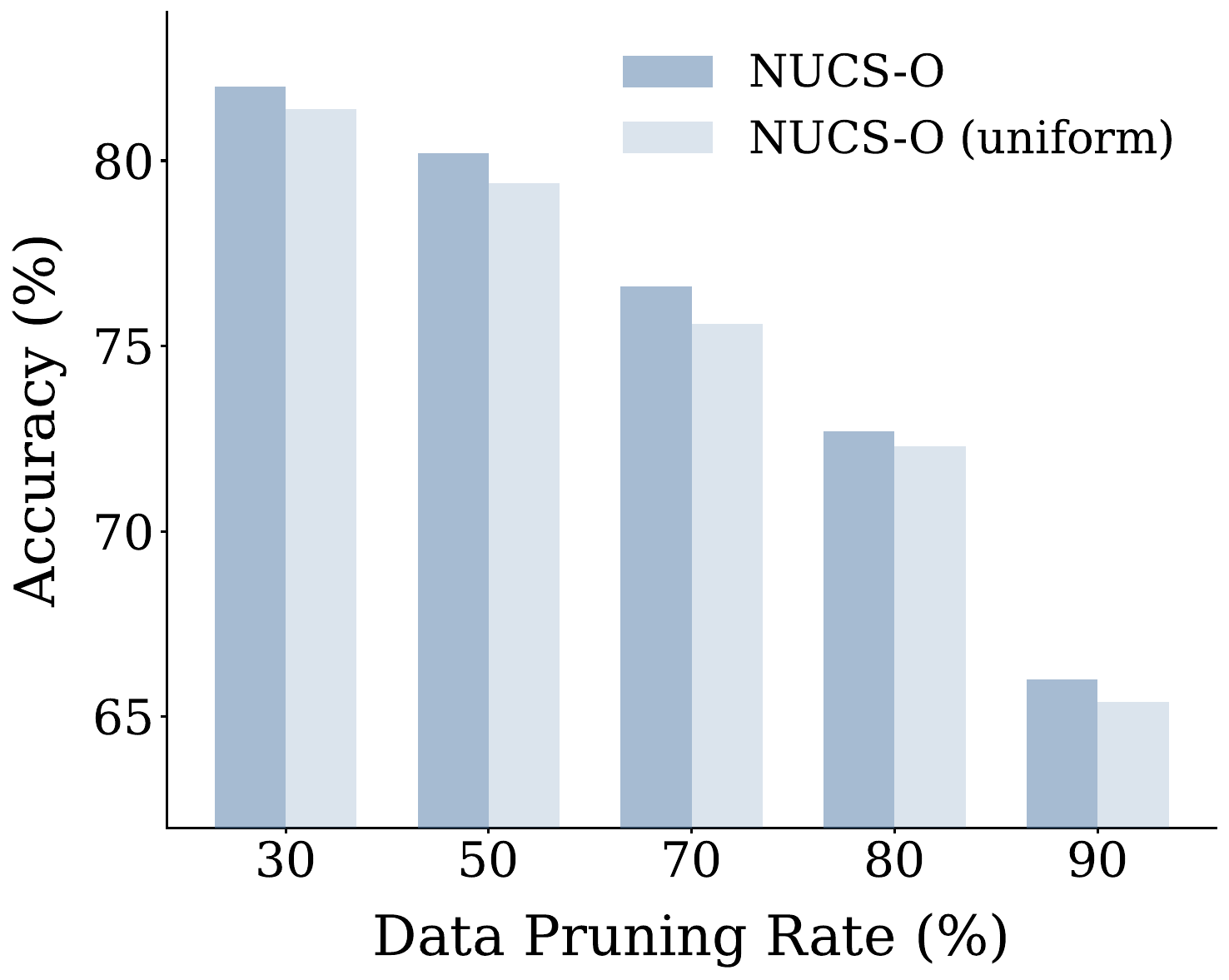} 
        \caption{}
        \label{fig:ab_class_num_subfig1}
    \end{subfigure}
    \begin{subfigure}[b]{0.48\columnwidth} 
        \centering
        \includegraphics[width=\textwidth, height=6cm, keepaspectratio]{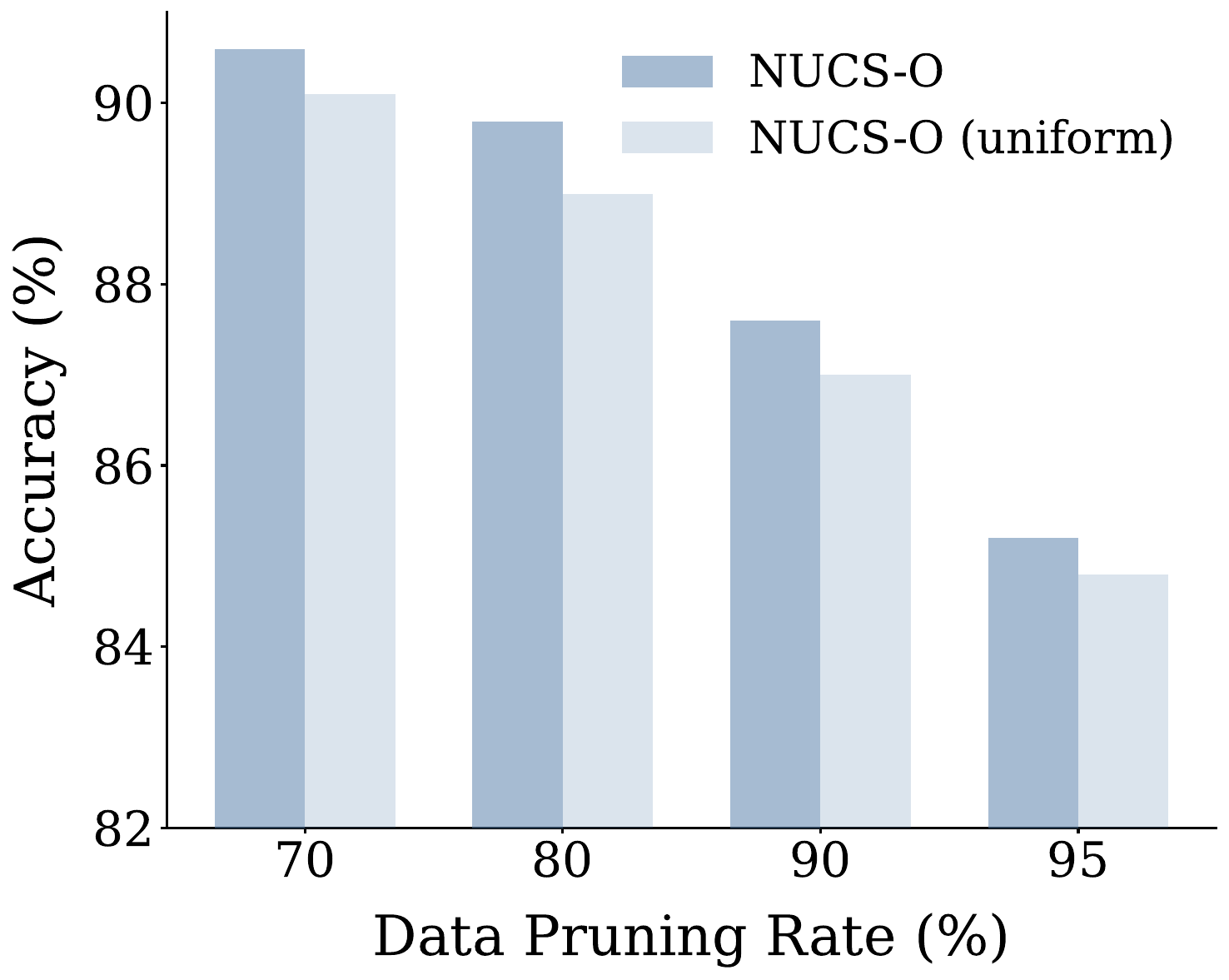} 
        \caption{}
        \label{fig:ab_class_num_subfig2}
    \end{subfigure}
    \vspace{-0.5pt} 
    \begin{subfigure}[b]{0.48\columnwidth} 
        \centering
        \includegraphics[width=\textwidth, height=6cm, keepaspectratio]{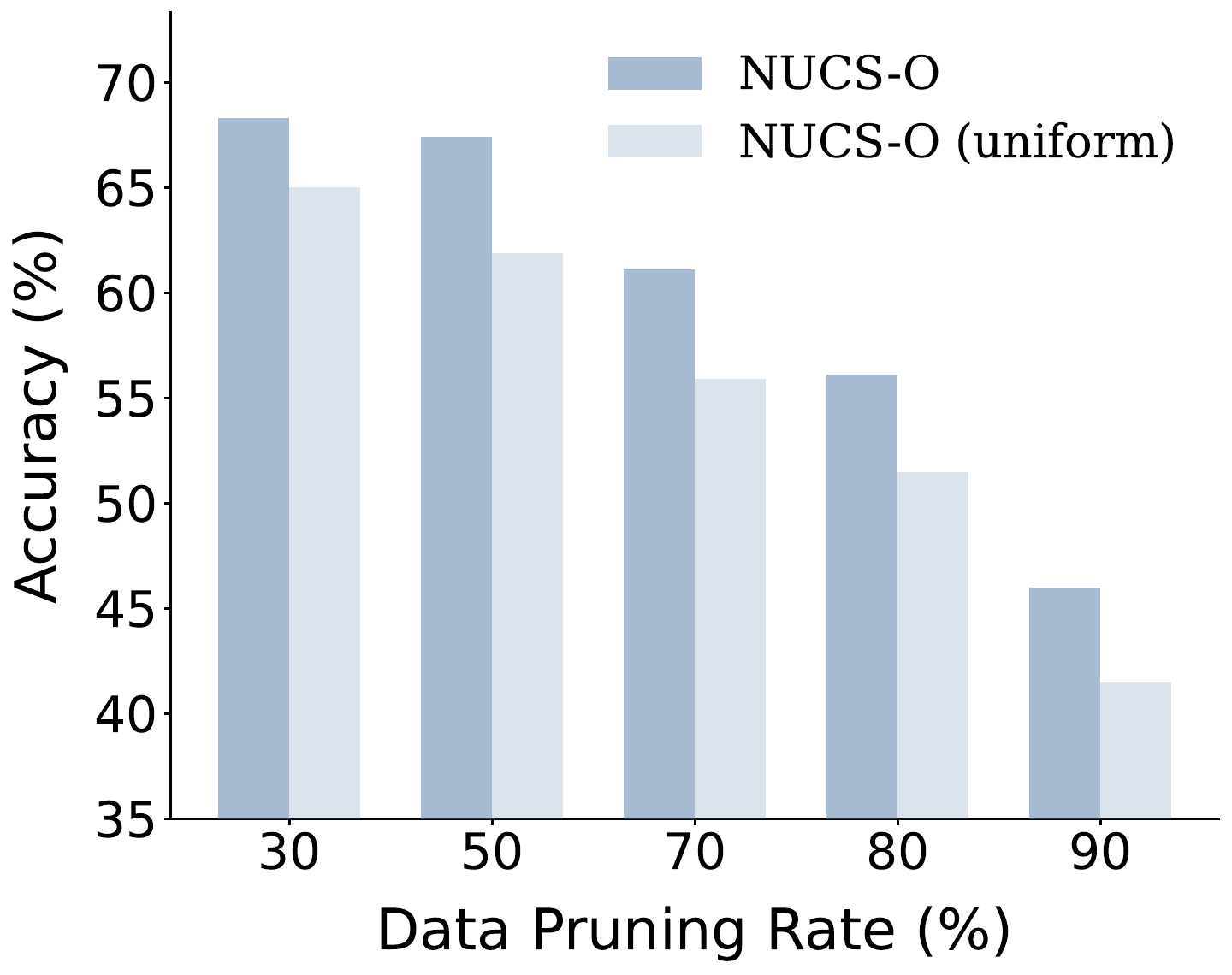} 
        \caption{}
        \label{fig:ab_class_num_subfig3}
    \end{subfigure}
    \begin{subfigure}[b]{0.48\columnwidth} 
        \centering
        \includegraphics[width=\textwidth, height=6cm, keepaspectratio]{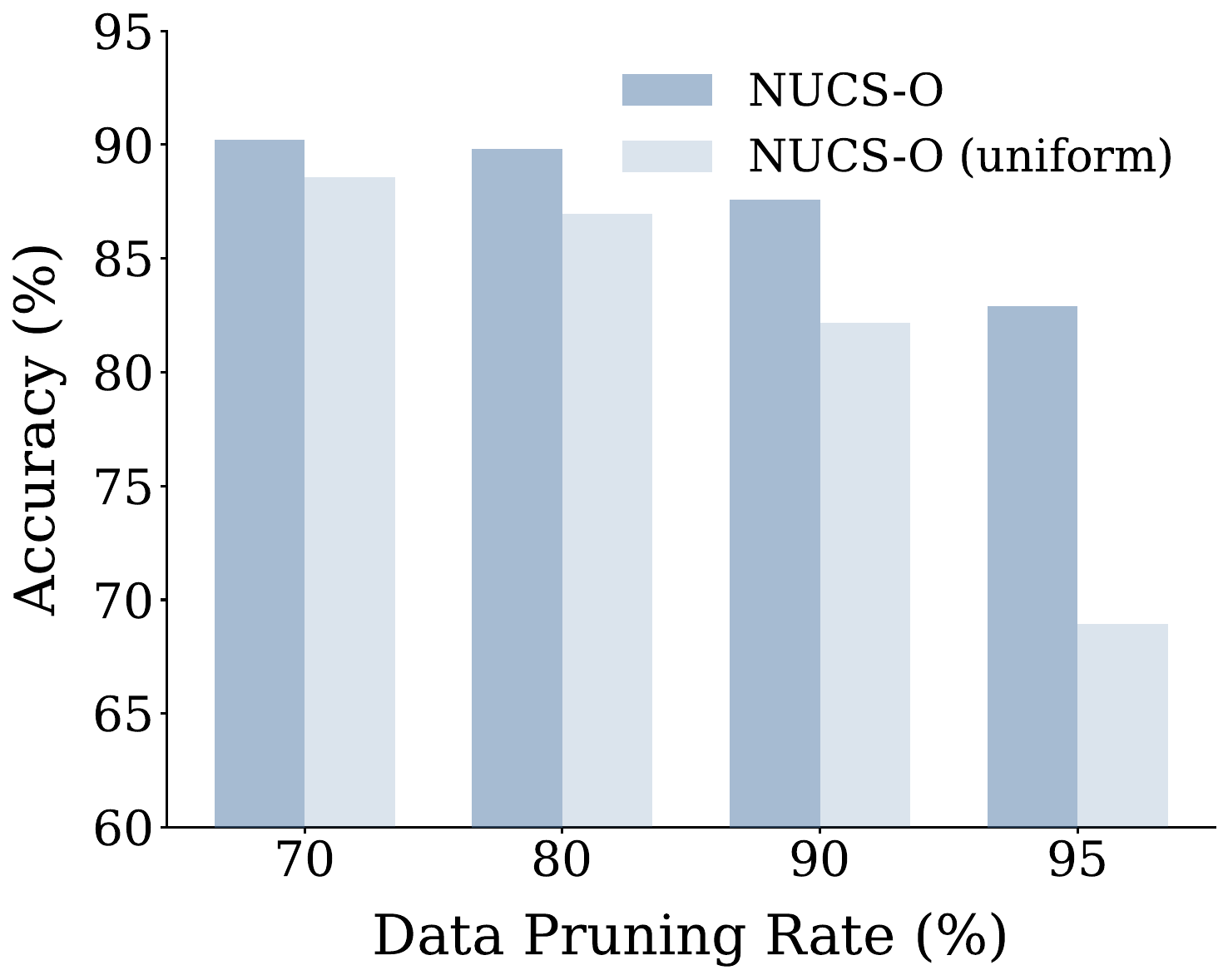} 
        \caption{}
        \label{fig:ab_class_num_subfig4}
    \end{subfigure}
    \caption{Ablation study comparing NUCS-O with its uniform budget allocation variant NUCS-O (uniform) on Food101 (a-b) and CIFAR100-LT (c-d) with ResNet18/ViT-L backbones.}
    \label{fig:ab_class_num}
\end{figure}

\begin{table*}[t]
\centering
\renewcommand{\arraystretch}{0.8}
\caption{Accuracy comparison of NUCS-O with its uniform budget allocation variant and class-specific method CCS-CP on CIFAR100-LT (I=20) and Food101 with ResNet18 (IN-1K) under different data difficulty metrics.}
\label{tab:combined_swapped}

\begin{NiceTabular}{llcccccccccccc}
\CodeBefore
  \rowcolor{gray!20}{3} 
\Body
\toprule
& Dataset ($\rightarrow$) & \multicolumn{6}{c}{\textbf{CIFAR100-LT}} & \multicolumn{6}{c}{\textbf{Food101}} \\
\cmidrule(lr){3-8} \cmidrule(lr){9-14}
& Pruning Rate ($\rightarrow$) & 0\% & 30\% & 50\% & 70\% & 80\% & 90\% & 0\% & 30\% & 50\% & 70\% & 80\% & 90\% \\
\midrule
Metric & Random & 67.9 & 64.7 & 60.9 & 54.3 & 47.1 & 36.6 & 82.8 & 80.9 & 78.7 & 75.1 & 71.7 & 64.8 \\
\midrule
\multirow{3}{*}{EL2N} & CCS-CP & - & 64.8 & 61.7 & 56.7 & 51.6 & 40.4 & - & 80.5 & 78.3 & 74.4 & 71.0 & 64.8 \\
& NUCS-O (uniform) & - & 65.0 & 61.9 & 55.9 & 51.5 & 41.5 & - & 81.4 & 79.4 & 75.6 & 72.3 & 65.4 \\
& NUCS-O & - & \textbf{67.9} & \textbf{67.4} & \textbf{61.1} & \textbf{56.1} & \textbf{46.0} & - & \textbf{82.0} & \textbf{80.2} & \textbf{76.6} & \textbf{72.7} & \textbf{66.0} \\
\midrule
\multirow{3}{*}{Effort} & CCS-CP & - & 65.5 & 62.6 & 57.4 & 51.5 & 40.9 & - & 81.0 & 78.8 & \textbf{75.2} & \textbf{72.0} & \textbf{65.6} \\
& NUCS-O (uniform) & - & 64.9 & 61.8 & 56.6 & 52.1 & 42.0 & - & 81.5 & 79.0 & 75.1 & 71.4 & 64.9 \\
& NUCS-O & - & \textbf{67.7} & \textbf{66.7} & \textbf{61.0} & \textbf{56.5} & \textbf{45.6} & - & \textbf{81.8} & \textbf{79.4} & \textbf{75.2} & 71.8 & 64.9 \\
\midrule
\multirow{3}{*}{AUM} & CCS-CP & - & 65.1 & 62.1 & 56.4 & 51.8 & 42.1 & - & 81.1 & 78.7 & 75.8 & 72.7 & \textbf{66.3} \\
& NUCS-O (uniform) & - & 65.3 & 62.2 & 57.0 & 52.6 & 42.4 & - & 81.5 & 80.0 & 76.4 & 73.1 & \textbf{66.3} \\
& NUCS-O & - & \textbf{68.5} & \textbf{67.5} & \textbf{62.7} & \textbf{57.9} & \textbf{48.1} & - & \textbf{82.2} & \textbf{80.6} & \textbf{77.0} & \textbf{73.4} & 65.9 \\
\bottomrule
\end{NiceTabular}
\end{table*}

\subsection{Ablation Study \& Discuss }
\label{sec:ablation}

\vspace{2pt}
\noindent
\textbf{Effectiveness of non-uniform sample strategy. }To further validate our difficulty-based class budget allocation strategy, we conduct experiments comparing NUCS-O with NUCS-O (uniform), where the latter strictly enforces uniform selection rates across all classes. Here, we use NUCS-O (rather than NUCS) to avoid potential biases introduced by linear ridge regression errors. Experiments are performed on Food101 and CIFAR100-LT with ResNet18 (IN-1K) and ViT-L (IN-21K). As shown in ~\Cref{fig:ab_class_num}, NUCS-O consistently outperforms NUCS-O (uniform) across different pruning rates and pre-trained models. These results highlight the effectiveness of our non-uniform allocation strategy. 



\vspace{2pt}
\noindent
\textbf{Evaluation across different data difficulty metrics. }To comprehensively evaluate the effectiveness of our method and non-uniform budget allocation strategy under different data difficulty metrics, we conduct experiments using the Effort and AUM score \cite{data,aum} on CIFAR100-LT and Food101 datasets. Notably, the original AUM metric incorporates negative values where lower scores indicate greater sample difficulty, we perform scale normalization by inverting the score polarity and enforcing value positivity. ~\Cref{tab:combined_swapped} illustrate the consistent performance advantage of our method when employing the proposed budget allocation strategy, as measured by both Effort and AUM metrics. This consistency can be evidenced by the intrinsic correlation patterns among different difficulty metrics \cite{sorscher}. While AUM demonstrates marginally superior performance, we note this comes at substantial computational cost - requiring complete fine-tuning epochs on downstream data. Through evaluation of this accuracy-efficiency trade-off, we ultimately adopt EL2N as our primary data difficulty metric.

\begin{table}[t!]
  \centering
  \caption{
  Comparison of Random, NUCS, and CCS-CP pruning methods on classification bias using Food101 at a 90\% pruning rate. Results are shown for ResNet18 and ViT-L backbones. Lower Diff. and higher WCA are better.}
  \label{tab:bias_compare_nice_fixed}
  \renewcommand{\arraystretch}{0.9} 
  \begin{NiceTabular}{@{}l S[table-format=2.1] S[table-format=1.2] S[table-format=2.1] S[table-format=1.2]@{}}
    \CodeBefore
      \rowcolor{gray!20}{3} 
    \Body
    \toprule 
    \Block{2-1}{\textbf{Method}} & \Block[c]{1-2}{\textbf{ResNet18}} & & \Block[c]{1-2}{\textbf{ViT-L}} \\
    \cmidrule(l){2-3} \cmidrule(l){4-5}
                          & {WCA (\%) $\uparrow$} & {Diff. $\downarrow$} & {WCA (\%) $\uparrow$} & {Diff. $\downarrow$} \\
    \midrule
    Random                & 20.4                  & 0.80                 & 56.4                  & 0.44                 \\ 
    CCS-CP                & 29.6                  & 0.69                 & 46.8                  & 0.53                 \\
    NUCS                  & \textbf{35.6}              & \textbf{0.58}             & \textbf{57.6}              & \textbf{0.42}             \\
    \bottomrule
  \end{NiceTabular}
\end{table}

\vspace{2pt}
\noindent
\textbf{Classification bias analysis. }The impact of coreset selection on classification bias has recently garnered significant attention \cite{drop}. In this work, we experimentally demonstrate that our proposed NUCS method not only achieves higher overall accuracy but also mitigates classification bias. We quantify this bias using two key metrics: (1) worst-class accuracy (WCA), reflecting performance on minority classes, and (2) the difference between maximum and minimum recall (Diff.), quantifying performance uniformity. As illustrated in \Cref{tab:bias_compare_nice_fixed}, NUCS exhibits lower bias than random selection and CCS-CP across both metrics.

\vspace{2pt}
\noindent
\textbf{Time efficiency comparison. }We evaluate the computational efficiency of NUCS, with a focus on its scalability to large-scale data. 
As detailed in Table~\ref{tab:time_combined_optimized}, our method delivers a significant 2$\times$ speedup on Food101 at 70\% pruning rate. Crucially, this substantial efficiency gain is maintained on the much larger iNaturalist dataset. This result demonstrates the robustness and effective scalability of our approach, confirming its practicality for resource-intensive applications.



\begin{table}[tbp]
\centering
\caption{
Time efficiency comparison on Food101 (75K samples) and the larger iNaturalist 2021 mini (500K samples). Experiments were conducted with ViT-L (IN-21K) on a single NVIDIA RTX 4090 GPU.
}
\renewcommand{\arraystretch}{0.9} 
\label{tab:time_combined_optimized}
\NiceMatrixOptions{cell-space-top-limit=2pt, cell-space-bottom-limit=2pt}

\begin{NiceTabular}{ll rr}
  \toprule
  \textbf{Dataset} & \textbf{Method} & Full Dataset & 30\% Subset \\
  \midrule
  \Block{2-1}{Food101} & Baseline & 2.6h & {-}  \\
                                 & NUCS    & {-}  & \textbf{1.3h} \\
  \midrule
  \Block{2-1}{iNaturalist} & Baseline & 23.3h & {-}  \\
                                    & NUCS     & {-}   & \textbf{11.3h} \\
  \bottomrule
\end{NiceTabular}
\end{table}

\vspace{2pt}
\noindent
\textbf{Cross-domain experiments. }To evaluate our method's performance in cross-domain settings, we select the NIH ChestX-ray14 dataset \cite{nih} as it exemplifies key challenges in medical imaging. The data's inherent difficulty presents a notable class clustering characteristic \cite{tsai2025class} and a severe class imbalance (imbalance factor $I=258$), both of which are common issues in medical datasets. Our results show that our method maintains a strong advantage even when facing this combination of complex conditions (details in \Cref{domain}). 





\section{Conclusion}
In this paper, we identify and analyze the limitations of existing coreset selection methods in overlooking critical class-level information and substantial inter-class difficulty variations. To address these issues, we propose Non-Uniform Class-Wise Coreset Selection (NUCS), a novel strategy that automatically determines class-specific data selection budgets based on global class difficulty and adaptively selects samples within optimal difficulty ranges in each class. Through comprehensive evaluations across diverse datasets and pre-trained models, we demonstrate the superior performance of our proposed pruning strategy. 
{
    \small
    \bibliographystyle{ieeenat_fullname}
    \bibliography{main}
}
\clearpage
\setcounter{page}{1}
\maketitlesupplementary

\appendix

\section{Limitation of Global Pruning Methods}
\label{difficultyharms}

\begin{figure}[t!]
    \centering 
    \includegraphics[width=0.8\columnwidth]{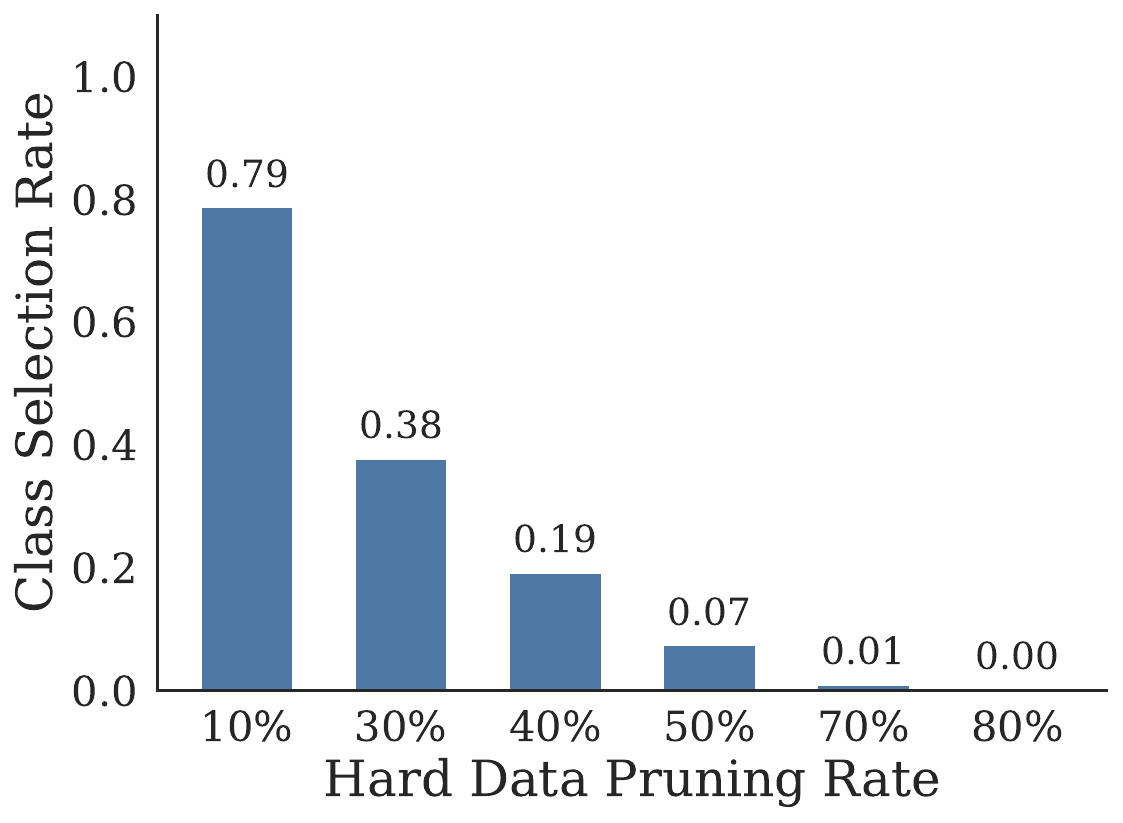}
    
    \caption{Data selection rate of 'applepie' class at different hard pruning rates. The results show global hard pruning strategy disproportionately prune certain class.} 
    \label{fig:my_example} 
\end{figure}

Our analysis reveals that a global pruning strategy such as classical hard data pruning can disproportionately impact certain classes. As illustrated in \Cref{fig:my_example}, for the 'applepie' class in the Food101 dataset, the data selection rate drops sharply when applying global hard data pruning to a ResNet18 model pre-trained on IN-1k. Specifically, at a 50\% global pruning rate, only 7\% of the class's data remains, plummeting to a mere 1\% at a 70\% pruning rate.

\section{EL2N Score Calculation Details}
\label{el2n}

EL2N \cite{el2n} has been demonstrated as a robust and efficient metric for quantifying data difficulty, requiring only a few training epochs for reliable estimation. For a specific sample $(x_i, y_i)$, where $y_i$ is the one-hot ground-truth label, its EL2N score $s_i$ is formally defined as the expected error over the course of training:
\begin{equation}
    s_{i} = \mathbb{E}_{t} \left\| p(x_{i}) - y_{i} \right\|_2,
    \label{eq:el2n_sample} 
\end{equation}
where $p( x_i)$ is the model's predicted probability distribution for input $x_i$ at a given time $t$.

In practice, we compute the EL2N scores for the IN-1k pre-trained ResNet18 and IN-21k pre-trained ViT-L models over the first 4 and 3 training epochs, respectively.

\section{Variation in Global Difficulty across Classes}
\label{cv of class}

To quantify the disparity in global difficulty variations across classes, we employ the Coefficient of Variation (CV), a standardized measure of dispersion and show result in two balanced datasets CIFAR100 and Food101. 
Given a set of global class difficulty scores $\mathcal{D} = \{S_1, S_2, \dots, S_Y\}$ for all $Y$ classes in a dataset, the CV is defined as the ratio of the standard deviation of these scores to their mean:
\begin{equation}
    \text{CV} = \frac{S_d}{\bar{S}},
    \label{eq:cv_definition}
\end{equation}
where $\bar{S}$ is the mean class difficulty and $S_d$ is the sample standard deviation. These components are calculated as follows:
\begin{align}
    \bar{d} &= \frac{1}{Y} \sum_{i=1}^{C} S_i, \label{eq:mean_difficulty} \\
    s_d &= \sqrt{\frac{1}{Y-1} \sum_{i=1}^{C} (S_i - \bar{S})^2}. \label{eq:std_difficulty}
\end{align}
A higher CV value signifies greater variance in difficulty across classes, indicating that the model finds some classes disproportionately harder or easier to learn than others. Conversely, a lower CV suggests a more uniform difficulty distribution. As shown in \Cref{tab:cv_difficulty}, the disparity in class difficulty is markedly more pronounced in pre-trained models than in those trained from scratch.

\begin{table}[t!]
    \centering 
    
    \caption{
        This table shows the coefficient of variation (CV) of global class difficulty scores. 
        The term 'ResNet18 Scratch' refers to a ResNet18 model trained from scratch. 
        The pre-trained models are a ResNet18 trained on ImageNet-1K and a ViT-L trained on ImageNet-21K.
    }
    \label{tab:cv_difficulty}
    
    \begin{NiceTabular}{l S[table-format=1.4] S[table-format=1.4]}
        
        \toprule 
        
        Model & {CIFAR100} & {Food101} \\ 
        
        \midrule 
        
        ResNet18 Scratch  & 0.1718 & 0.1485 \\
        ResNet18 Pre-trained & 0.3014 & 0.1997 \\
        ViT-L Pre-trained   & 0.5682 & 0.3497 \\
        
        \bottomrule 
        
    \end{NiceTabular}
\end{table}

\section{Detailed Experimental Setting}
\label{app:setting}

\subsection{Dataset Benchmarks}
We utilize four balanced benchmark datasets in our experiments: CIFAR10 \cite{cifar}, CIFAR100, Food101 \cite{food}, and iNaturalist 2021 Mini \cite{inaturalist2021}. The CIFAR10 dataset contains 60,000 color images evenly distributed across 10 classes, with each class comprising 6,000 images. The dataset is split into 50,000 training images and 10,000 test images. CIFAR100 extends this to 100 fine-grained classes, with each class containing 500 training images and 100 test images. Food101 consists of 101 food categories with a total of 101,000 images. Each class includes 750 training images and 250 manually verified test images. For the iNaturalist 2021 dataset, we use the mini training version which covers 10,000 species. This subset contains 50 training examples per species, totaling 500,000 images. We follow the official training and validation splits provided by each dataset in all our experiments.

In our experiments, we employ long-tailed versions of CIFAR100 with imbalance factors I=20. The imbalanced CIFAR100 datasets are created following the approach of \cite{classbalance} by subsampling each class $ k \in [K] $ to retain $ \mu^{k-1} $ of its original size. For an initially balanced dataset, this results in an imbalance factor $ I = \mu^{1-K} $, representing the size rate between the largest and smallest classes. Additionally, we evaluate our approach on the NIH ChestX-ray14 dataset. For the NIH dataset, we follow the data processing pipeline described in \cite{nihlimits}, resulting in approximately 30,000 highly imbalanced images. To properly account for label imbalance in NIH, we use the Area Under the Curve (AUC) metric to evaluate model performance.

\subsection{Training Details}


\begin{table}[t]
  \centering
  \caption{
    \textbf{Fine-tuning Hyperparameters.} 
    Summary of fine-tuning hyperparameters for ResNet18 (pretrained on ImageNet-1k) and ViT-Large (pretrained on ImageNet-21k) across all datasets. 
    The settings for the standard and long-tailed (LT) versions of the CIFAR datasets are identical.
  }
  \label{tab:finetuning_hyperparams_simplified}
  \begin{tabular}{@{}llcc@{}}
    \toprule
    \textbf{Dataset} & \textbf{Model} & \textbf{Epochs} & \textbf{Batch Size} \\
    \midrule
    \multirow{2}{*}{CIFAR10} & ResNet18 & 30 & 64 \\
    & ViT-L & 15 & 16 \\
    \addlinespace 
    \multirow{2}{*}{CIFAR100 (LT)} & ResNet18 & 30 & 64 \\
    & ViT-L & 15 & 16 \\
    \addlinespace
    \multirow{2}{*}{Food101} & ResNet18 & 30 & 256 \\
    & ViT-L & 15 & 32 \\
    \addlinespace
    \multirow{2}{*}{iNaturalist 2021 Mini} & ResNet18 & 40 & 256 \\
    & ViT-L & 20 & 32 \\
    \bottomrule
  \end{tabular}
\end{table}

In our experiments, we train the ViT-Large model using the parameter-efficient LoRA technique \cite{hu2022lora}. After conducting a grid search for optimal learning rates, we set the learning rate to $5e^{-3}$ for both ResNet18 (pretrained on IN-1K) and ViT-Large (pretrained on IN-21K).  We use the SGD optimizer with a momentum of 0.9 and weight decay of 0.0005. The learning rate is scheduled using the cosine annealing strategy \cite{loshchilov2022sgdr}, with a minimum learning rate of 0.0001. Note that because of huge calculation consumptions, the experiment in each case is performed once. The fine-tuning hyperparameters are shown in \Cref{tab:finetuning_hyperparams_simplified}.

For each dataset, we list the optimal window fraction endpoint $\beta$ for every data pruning rate $\alpha$ in the format of $(\alpha, \beta)$.

\noindent ResNet18 (IN-1K):
\begin{itemize}
    \setlength\itemsep{0em}
    \item CIFAR10: (60, 1.0), (70, 1.0), (80, 1.0), (90, 0.9).
    \item CIFAR100: (30, 1.0), (50, 1.0), (70, 0.8), (80, 0.7), (90, 0.6).
    \item Food101: (30, 1.0), (50, 0.9), (70, 0.8), (80, 0.7), (90, 0.6).
    \item CIFAR100-LT: (30, 1.0), (50, 0.8), (70, 0.7), (80, 0.4), (90, 0.1).
\end{itemize}

\noindent ViT-L (IN-21K):
\begin{itemize}
    \setlength\itemsep{0em}
    \item CIFAR10: (90, 1.0), (95, 0.9), (98, 0.9), (99, 0.6).
    \item CIFAR100: (70, 1.0), (80, 0.9), (90, 0.8), (95, 0.8).
    \item Food101: (70, 1.0), (80, 0.9), (90, 0.8), (95, 0.7).
    \item CIFAR100-LT: (70, 1.0), (80, 0.9), (90, 0.7), (95, 0.4).
\end{itemize}


\section{Evaluation Result}
\label{app:result}

\subsection{Evaluation on NIH Dataset}
\label{domain}

To evaluate our method's performance in cross-domain settings, we report our method's performance on NIH ChestX-ray14 dataset \Cref{tab:nih_results_baseline}. The NIH dataset is a multi-label classification task where the class difficulty distribution exhibits a distinct clustering, a common trait in medical imaging. In this challenging scenario, we find that methods like CCS, which perform well on natural images, can surprisingly yield results worse than Random Pruning. In contrast, NUCS consistently outperforms this random baseline, with NUCS-O consistently achieving the best performance.

\colorlet{mygray}{gray!20}

\begin{table}[t!]
  \centering
  \caption{Comparison of methods' performance (AUC) on the NIH dataset under different data pruning rates. The model is a ResNet34 pre-trained on ImageNet-1K.}
  \label{tab:nih_results_baseline}
   \setlength{\tabcolsep}{4pt}
  \begin{NiceTabular}{lcccccc}
    \CodeBefore
      \rowcolor{mygray}{3}
    \Body
    \toprule
    & \Block{1-6}{Pruning rate} \\
    \cmidrule(l){2-7} 
    \textbf{Method} & \textbf{0\%} & \textbf{30\%} & \textbf{50\%} & \textbf{70\%} & \textbf{80\%} & \textbf{90\%} \\
    \midrule
    Random & 0.802 &0.794 & 0.778 & 0.752 & 0.740 & 0.730 \\
    EL2N & - &0.769 & 0.737 & 0.736 & 0.733 & 0.696 \\
    CCS & - &0.782 & 0.772 & 0.748 & 0.736 & 0.722 \\
    D2 & - & \textbf{0.803} & 0.782 & 0.766 & 0.750 & 0.713 \\
    BWS & - & 0.788 & 0.777 & 0.761 & 0.754 & 0.737 \\
    CCS-CP & - & 0.776 & 0.780 & 0.769 & 0.763 & 0.753 \\
    \midrule
    NUCS & - & \textbf{0.803} & \textbf{0.798} & 0.754 & 0.766 & 0.732 \\
    NUCS-O & - & \textbf{0.803} & \textbf{0.798} & \textbf{0.774} & \textbf{0.769} & \textbf{0.756} \\
    \bottomrule
  \end{NiceTabular}
\end{table}

\subsection{CIFAR10 Results}

Here we provide the evaluation results on CIFAR10 dataset with ResNet18 (IN-1K) pre-trained and ViT-L (IN-21K) pre-trained models. Full results are provided in \Cref{tab:combined_results}.

\begin{table*}[t!]
  \centering
  \caption{Comparison of pruning methods on CIFAR-10, fine-tuned on ResNet-18 and ViT-L models.}
  \label{tab:combined_results}
  \begin{NiceTabular}{lcccccccccc}
  \CodeBefore
      \rowcolor{mygray}{3}
  \Body
    \toprule
    \Block{2-1}{\textbf{Method}} & \Block{1-5}{ResNet-18 (IN-1K)} & & & & & \Block{1-5}{ViT-L (IN-21K)} \\
    \cmidrule(lr){2-6} \cmidrule(lr){7-11}
    & \textbf{0\%} & \textbf{60\%} & \textbf{70\%} & \textbf{80\%} & \textbf{90\%} & \textbf{0\%} & \textbf{90\%} & \textbf{95\%} & \textbf{98\%} & \textbf{99\%} \\
    \midrule
    Random     & 96.7 & 95.3          & 94.9          & 93.9          & 92.2 & 98.9 & 98.7 & 98.5 & 98.3 & 97.8 \\
    EL2N       & -    & 96.6          & \textbf{96.4} & 95.6          & 88.8 & -    & 98.9 & 98.8 & 62.1 & 31.9 \\
    Moderate   & -    & 95.4          & 95.1          & 94.2          & 92.0 & -    & 98.5 & 98.4 & 98.1 & 97.5 \\
    CCS        & -    & \textbf{96.7} & 96.0          & 95.5          & 93.3 & -    & \textbf{99.0} & 98.8 & 98.4 & 98.1 \\
    D2         & -    & 96.6          & 96.3          & 95.5          & 93.2 & -    & 98.9 & 98.8 & 90.5 & 51.8 \\
    BWS        & -    & 96.6          & 96.2          & 94.9          & 93.1 & -    & 98.8 & 98.7 & 98.1 & 97.9 \\
    CCS-CP     & -    & 96.3          & 96.1          & 95.2          & 93.4 & -    & \textbf{99.0} & 98.7 & \textbf{98.5} & \textbf{98.2} \\
    \midrule
    NUCS       & -    & 96.5          & 96.3          & 95.1          & 92.4 & -    & \textbf{99.0} &\textbf{98.9} & 98.1 & 97.9 \\
    NUCS-O     & -    & 96.5          & 96.3          & \textbf{95.7} & \textbf{93.9} & - & \textbf{99.0} & \textbf{98.9} & \textbf{98.5} & 98.0 \\
    \bottomrule
  \end{NiceTabular}
\end{table*}

\section{More illustration about Theoretical Analysis.}

For simplicity, the main text derives the optimal class selection rate by setting the partial derivatives of the error function $E(t, f_0)$ to zero, initially ignoring boundary constraints. Here we provide a complete analysis that account for the coreset constraints on the selection rates $f_0$ and $f_1$. The optimization problem is more formally stated as:
\begin{equation}
\begin{aligned}
& \min_{t, f_0} & & E(t, f_0) = f_0 \Phi\left(\frac{\mu_0-t}{\sigma_0}\right)+f_1\Phi\left(\frac{t-\mu_1}{\sigma_1}\right) \\
& \text{subject to} & & f_0 + f_1 = 2f, 
 0 \le f_0, f_1 \le 1.
\end{aligned}
\label{eq:constrained_opt}
\end{equation}

The unconstrained solution, derived by ignoring the inequality constraints $f_i \le 1$, yields the ideal allocation rates, which we denote by $(f_0^*, f_1^*)$:
\begin{align}
    f_0^* &= \frac{2f \sigma_0}{\sigma_0 + \sigma_1}, \\
    f_1^* &= \frac{2f \sigma_1}{\sigma_0 + \sigma_1}.
\end{align}
This solution is only valid if it lies within the feasible region defined by the constraints in Eq.~\eqref{eq:constrained_opt}.

A boundary condition occurs when the relative difficulty of one class, as measured by its variance $\sigma_i$ in the toy example, is so high that the unconstrained solution suggests allocating more than 100\% of its available data. This happens when either $f_0^* > 1$ or $f_1^* > 1$. Consider the case where $f_0^* > 1$, which implies $\sigma_0(2f-1) > \sigma_1$. This indicates that class $\mathcal{D}_0$ is the dominant source of error, and the model attempts to over-allocate resources to it. Since the objective function $E(t, f_0)$ is convex with respect to $f_0$ (for a fixed optimal $t$), the minimum within the constrained interval must lie at the boundary closest to the unconstrained minimum $f_0^*$. Therefore, the optimal selection rate for class $\mathcal{D}_0$ becomes saturated at its upper bound. The optimal allocation strategy is thus satisfying:
\begin{equation}
    f_0^{\text{opt}} = 1, \quad f_1^{\text{opt}} = 2f - 1.
\end{equation}

By symmetry, if class $\mathcal{D}_1$ is disproportionately difficult such that $f_1^* > 1$, the optimal allocation is:
\begin{equation}
    f_1^{\text{opt}} = 1,  \quad f_0^{\text{opt}} = 2f - 1.
\end{equation}

We can summarize the complete, constrained optimal allocation strategy $(f_0^{\text{opt}}, f_1^{\text{opt}})$ as a piecewise function. The optimal rate for class $\mathcal{D}_0$ is given by:
\begin{equation}
f_0^{\text{opt}} = 
\begin{cases}
    1 & \text{if } \frac{2f \sigma_0}{\sigma_0 + \sigma_1} > 1 \\
    2f - 1 & \text{if } \frac{2f \sigma_1}{\sigma_0 + \sigma_1} > 1 \\
    \frac{2f \sigma_0}{\sigma_0 + \sigma_1} & \text{otherwise}
\end{cases}
\label{eq:piecewise_f0}
\end{equation}
and the rate for class $\mathcal{D}_1$ is determined by the budget constraint, $f_1^{\text{opt}} = 2f - f_0^{\text{opt}}$.

This analysis offers a refinement to our core hypothesis and directly informs the design of our algorithm for extreme scenarios. While the optimal data allocation is generally proportional to class difficulty ($\sigma_i$), this relationship holds for an interior solution. When the difficulty of one class becomes a significant bottleneck, the optimal strategy shifts from proportional allocation to a priority-based approach, where the budget for the most challenging class is saturated at its maximum possible value before allocating the remainder to the easier class.

\end{document}